\documentclass[10pt,twocolumn,letterpaper]{article}

\usepackage[pagenumbers]{cvpr} 

\definecolor{cvprblue}{rgb}{0.21,0.49,0.74}
\usepackage[pagebackref,breaklinks,colorlinks,allcolors=cvprblue]{hyperref}
\usepackage{multirow}
\usepackage{amsmath,amsfonts,amssymb}
\usepackage{booktabs}
\usepackage{pifont}
\usepackage[table,svgnames]{xcolor}
\usepackage{colortbl}
\usepackage{pgf}
\usepackage{svg}
\svgsetup{inkscapelatex=false}
\newcommand{\cmark}{\ding{51}}%
\def\paperID{13427} 
\def\confName{CVPR}
\def\confYear{2026}

\title{Scaling Self-Supervised and Cross-Modal Pretraining for Volumetric CT Transformers}

\author{\textbf{Cris Claessens$^{*, \, 1}$} \and \textbf{Christiaan Viviers$^{*, \, 1}$} \and \textbf{Giacomo D'Amicantonio$^{2}$} \and \textbf{Egor Bondarev$^{2}$} \and \textbf{Fons van der Sommen$^{1}$}\\
Dept. of Electrical Engineering, Eindhoven University of Technology, Eindhoven, The Netherlands\\
$^{1}$ARIA Lab, $^{2}$AIMS Lab, $^{*}$Contributed equally\\
{\tt\small \{c.h.b.claessens, c.g.a.viviers, g.d.amicantonio, e.bondarev, fvdsommen\}@tue.nl}
}

\begin{document}
\maketitle

\begin{abstract}
    We introduce SPECTRE, a fully transformer-based foundation model for volumetric computed tomography~(CT). Our \underline{S}elf-Supervised \& Cross-Modal \underline{P}r\underline{e}training for \underline{CT} \underline{R}epresentation \underline{E}xtraction~(SPECTRE) approach utilizes scalable 3D Vision Transformer architectures and modern self-supervised and vision–language pretraining strategies to learn general-purpose CT representations. Volumetric CT poses unique challenges, such as extreme token scaling, geometric anisotropy, and weak or noisy clinical supervision, that make standard transformer and contrastive learning recipes ineffective out of the box. The framework jointly optimizes a local transformer for high-resolution volumetric feature extraction and a global transformer for whole-scan context modeling, making large-scale 3D attention computationally tractable. Notably, SPECTRE is trained exclusively on openly available CT datasets, demonstrating that high-performing, generalizable representations can be achieved without relying on private data. Pretraining combines DINO-style self-distillation with SigLIP-based vision–language alignment using paired radiology reports, yielding features that are both geometrically consistent and clinically meaningful. Across multiple CT benchmarks, SPECTRE consistently outperforms prior CT foundation models in both zero-shot and fine-tuned settings, establishing SPECTRE as a scalable, open, and fully transformer-based foundation model for 3D medical imaging.\footnote{Code available at: \url{https://github.com/cclaess/SPECTRE}}
\end{abstract}    
\section{Introduction}\label{sec: Introduction}
\begin{figure}[t!]
    \centering
    \includegraphics[width=0.90\linewidth]{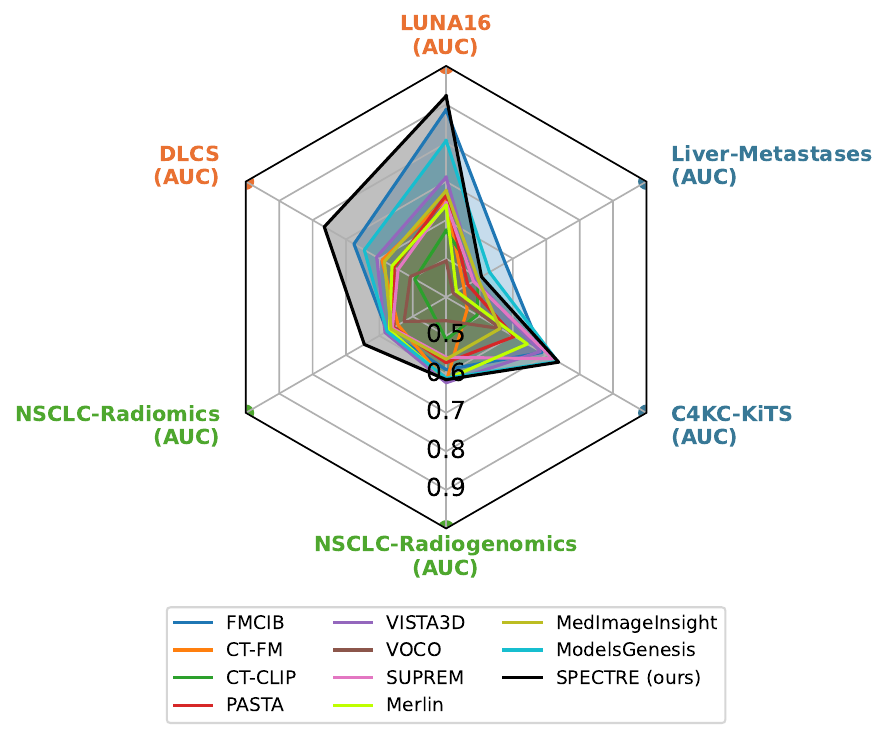}
    \caption{Radar plot comparing 11 CT foundation models across six biomarker classification benchmarks using frozen-embedding kNN classifiers. Diagnostic tasks on chest CT are shown in orange, prognostic tasks on chest CT in green, and prognostic tasks on abdominal CT in blue. SPECTRE achieves the highest performance on four of the six benchmarks, demonstrating stronger and more transferable volumetric representations compared to prior models.}
    \label{fig:radar_plot_foundation_models}
\end{figure}
Self-supervised learning~(SSL) and vision-language alignment~(VLA) are two rapidly maturing paradigms for learning high-quality visual representations in computer vision. SSL comprises methods that construct surrogate objectives from unlabeled images so that models learn invariances and mid-level features without any text or structured labels~\cite{chen_simple_2020, grill_bootstrap_2020}. In contrast, VLA directly couples visual encoders with textual encoders through alignment objectives so that learned features carry high-level compositional semantics grounded in language~\cite{radford_learning_2021}. These two families of objectives address different statistical problems: SSL provides dense, data-efficient priors about image structure, while VLA injects explicit semantic grounding that is essential for many downstream retrieval and reasoning tasks.

Several recent works have adapted elements of SSL and VLA to 2D medical imaging, yielding foundation models that transfer well across different downstream tasks~\cite{bannur_learning_2023, boers_foundation_2024, chen_towards_2024, jaspers_scaling_2025}. However, these successes do not automatically translate to volumetric clinical imaging, such as computed tomography~(CT). One reason is architectural: Vision Transformers~(ViTs) are the backbone of many modern foundation models because they provide a flexible, scalable attention mechanism with minimal hand-crafted inductive bias~\cite{dosovitskiy_image_2020}. In natural images, that flexibility is an advantage when large, diverse datasets and expressive pretraining objectives are available. In medical imaging, and especially in volumetric modalities, the absence of strong intrinsic locality and translational biases has historically slowed transformer adoption: in low-data or heterogeneous-data regimes, these priors act as regularizers that improve sample efficiency and robustness~\cite{dascoli_convit_2021}. In the context of CT imaging, these geometrical and local priors need to be introduced in the model by design, through tokenization, positional encodings, attention design, or auxiliary objectives rather than rely on these biases to be present by default.

Volumetric CT shifts the technical problem in several fundamental ways that interact directly with both architecture and objective choice. First, naive patching of 3D volumes produces token counts that scale roughly cubically with resolution, and transformer self-attention scales quadratically in token number~\cite{keles_computational_2022}. Some of the most effective design choices in transformers at scale, such as global attention or large training batches, require a  certain degree of approximation to be employed for CT imaging. Second, CT commonly exhibits strong geometric heterogeneity: anisotropic voxel spacing, variable field-of-view~(FOV), and scanner-specific preprocessing (reconstruction kernels, denoising, dose modulation). Positional encodings and receptive field parameterizations that assume isotropy or simple translation invariance will therefore misrepresent interscan geometry. Hence, models must explicitly represent voxel spacing and slice sampling or learn geometry-aware receptive fields to generalize across protocols~\cite{gayap_salm_2025}. Third, clinical supervision is typically weak, noisy, and hierarchical~(free-text reports, sparse tags, study-level labels), creating a tension between dense objectives that teach spatial precision (masked modeling, reconstruction) and global alignment objectives that teach semantic consistency with language (contrastive VLA losses)~\cite{hamamci_foundation_2024}. Finally, many 2D VLA recipes succeed because they can exploit huge negative sets and large effective batch sizes. In 3D those levers are severely limited by memory and computation. The problem is further amplified in medical imaging, as radiology reports and diagnostic codes often list multiple co-occurring conditions. Hence, candidate “negatives” frequently share clinical semantics with positives. This overlap weakens the signal from standard CLIP-style contrastive losses.

\begin{figure*}
    \centering
    \includegraphics[width=0.90\linewidth]{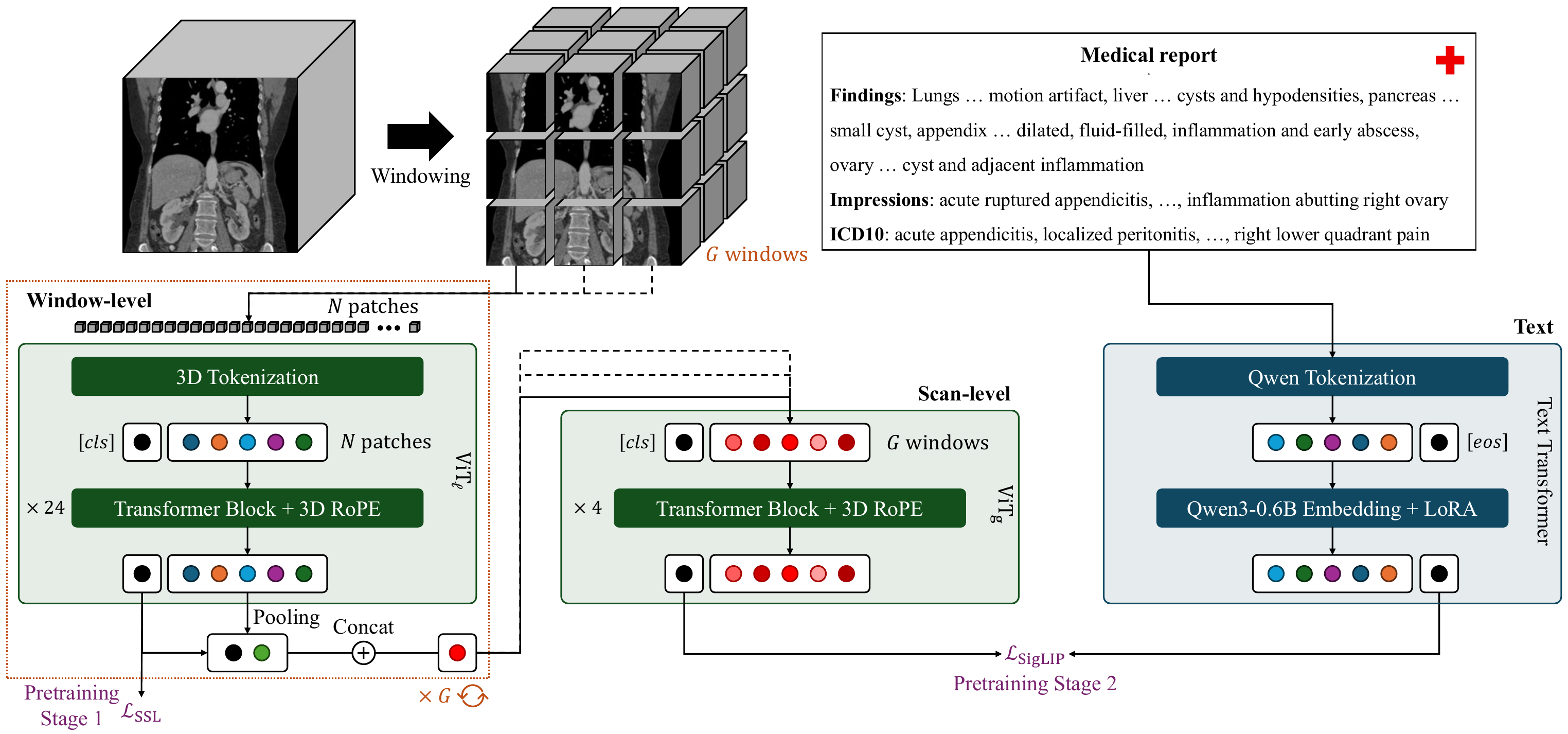}
    \caption{Overview of the proposed multimodal CT–report model. The model jointly processes volumetric CT data and corresponding radiology reports. The local vision transformer~ViT$_{\ell}$, pretrained using DINOv3~(Stage~1), extracts localized image features from CT volume crops. These features are aggregated by the global vision transformer~ViT$_g$, while the text transformer encodes the associated medical report. During SigLIP pretraining~(Stage~2), the vision and text representations are aligned in a shared embedding space.}
    \label{fig:overview}
\end{figure*}

These technical constraints motivate a set of open questions about how to build scalable, generalizable 3D foundation models. In this work, we treat these questions as the central technical problems rather than as secondary engineering constraints. Specifically, we introduce SPECTRE, a transformer-based 3D CT foundation model trained on industrial-grade hardware. Our framework emphasizes (1)~geometry-aware transformer design and tokenization to encode anisotropy and voxel-scale information explicitly, (2)~attention architectures and computational strategies that balance token complexity and context preservation for large-scale volumetric data, and (3)~a two-stage pretraining pipeline that combines SSL and VLA objectives; using SSL-like objectives to bootstrap robust geometry-aware features, followed by VLA to inject clinical semantics. We empirically analyze how these design choices affect downstream transfer across tasks that span region-level localization, classification, and study-level semantic retrieval. To support future research on scalable medical foundation models, we publicly release SPECTRE, along with all training code and pretraining recipes, as a fully open-source foundation for the community. An overview of the SPECTRE architecture and pretraining is presented in \cref{fig:overview}.
\section{Related Works}\label{sec: Related Works}

\subsection{3D Vision Transformers}
\label{sec: Related works / 3D ViT}
Recent transformer-based architectures have been extended to 3D medical volumes to capture global context beyond the reach of local convolutions. Early models embed a transformer encoder within a U‑shaped network: for example, UNETR~\cite{hatamizadeh_unetr_2022} uses a pure transformer backbone to encode the entire volume as a sequence, then connects multi-scale features via U‑Net skip-connections. Similarly, nnFormer~\cite{zhou_nnformer_2023} interleaves convolutional stems with transformer blocks and adds novel volume-based self-attention and “skip attention” for U‑Net connections. These hybrid designs exploit convolutions for local detail while leveraging self-attention for long-range dependencies.

Hierarchical and windowed attention schemes have also been popular. Swin-style 3D transformers embed patches hierarchically: SwinUNETR~\cite{crimi_swin_2022} uses 3D windowed attention to build a multi-resolution encoder-decoder architecture. The original SwinUNETR achieved SOTA on several benchmarks, and SwinUNETR-V2~\cite{greenspan_swinunetr-v2_2023} further inserts convolutional layers before each Swin block to reintroduce spatial bias, yielding a stronger backbone that generalizes across tasks with a single recipe. These models downsample tokens to form feature pyramids, enabling efficient computation on large volumes.

More recent pure-transformer architectures eliminate convolutions entirely and refine the fundamental ViT components for volumetric data. Primus~\cite{wald_primus_2025} introduces the first fully transformer 3D segmentation network, preserving high-resolution tokens and employing improved block designs with 3D rotary positional embeddings~(RoPE) to encode volumetric geometry. SuperFormer~\cite{forigua_superformer_2022} generalizes Swin Transformer~\cite{liu_swin_2021} to 3D super-resolution via volumetric patch embeddings, 3D relative positional encoding, and shifted window attention, whereas WaveFormer~\cite{gee_waveformer_2026} integrates multi-scale wavelet transforms inside the transformer to retain high-frequency detail with reduced complexity. Across these models, volumetric patch embeddings replace 2D patches with 3D cubes, while attention is commonly employed within local 3D windows or solely in the axial direction.

SPECTRE builds on these advances with a fully volumetric transformer backbone tailored to CT. It adopts anisotropic 3D patch embeddings aligned with CT voxel geometry and employs 3D RoPE for volumetric relative positioning, but extends it with DINOv3-style stochastic shifts~\cite{simeoni_dinov3_2025} during pretraining to increase robustness to variable voxel spacing and FOV. Further, SPECTRE uses a two-stage attention design combining dense local attention with coarse global attention, and is trained at foundation scale on a large unlabeled CT corpus, enabling general-purpose volumetric representation learning.

\subsection{CT Foundation Models}
\label{sec: Related works / CT FM}
In CT imaging, \emph{foundation models} leverage large unlabeled datasets or rich supervision. CT-CLIP~\cite{hamamci_foundation_2024} and Merlin~\cite{blankemeier_merlin_2024} align 3D CT with text. CT-CLIP uses $\approx25{,}000$ chest CTs and a CLIP-style loss~\cite{radford_learning_2021} for joint embeddings, excelling at retrieval and detection. Merlin trains on $\approx15{,}000$ abdominal CTs with reports and EHR labels, enabling strong zero-shot classification, retrieval, report generation, and segmentation. These VLA models are effective but region-specific. MedImageInsight~\cite{codella_medimageinsight_2024} takes a slightly different approach and scales to diverse modalities, including X-ray, CT, and MRI. However, it may under-emphasize fine CT anatomy compared to CT specific models.

A second family uses SSL on large unlabeled CT datasets. CT-FM~\cite{pai_vision_2025} pre-trains on 148,000~mixed volumes with contrastive objectives, achieving state-of-the-art segmentation, triage, retrieval, and semantic performance on several tasks while clustering anatomical regions. FMCIB~\cite{pai_foundation_2024} contrastively pre-trains on 11,467~lesion patches for biomarker prediction, outperforming ``from scratch" and ImageNet~\cite{deng_imagenet_2009} baselines and correlating with tumor biology. VoCo~\cite{wu_voco_2024} adds region-level context prediction by identifying anatomical location from ``base’’ crops. SSL models capture strong image features, however, they may lack explicit clinical semantics.

Some models target segmentation. VISTA3D~\cite{he_vista3d_2025} unifies supervised and interactive segmentation through a CNN encoder and promptable interface. SuPreM~\cite{li_how_2025}, built on aggregated labeled datasets~\cite{li_abdomenatlas_2024}, pre-trains supervised 3D encoders with high transfer performance. These models show excellent performance on segmentation tasks, but depend on dense labels and focus narrowly on anatomy.

PASTA~\cite{lei_synthetic_2025} takes a generative approach and uses 30,000~synthetic CTs with tumor annotations and reports to pre-train a model excelling on 45/46 downstream oncology tasks. Additionally, it builds a clinical decision-support system, improving radiologist accuracy. Synthetic pipelines reduce data scarcity but risk simulation bias and task specialization.

Earlier work like ModelsGenesis~\cite{zhou_models_2021} showed that simple 3D SSL~(\ie, inpainting, shuffling) outperforms training from scratch and 2D ImageNet transfer, but remains small-scale and unimodal.

In summary, CT foundation models vary in modality (image-only vs. multimodal), body coverage, and pretraining strategy. Most VLA models target one region and one modality, while vision-only SSL models focus on image features without language. Segmentation models emphasize per-voxel labels. SPECTRE integrates these strengths: multi-region CT pretraining, a pure 3D ViT with explicit geometry encoding, and a two-stage pipeline that first learns robust SSL-like primitives and then adds clinical semantics via VLA on report-paired data, combining fine 3D detail with broad clinical understanding in an open-source, scalable framework.
\section{Efficient 3D Transformer-Based Modeling}\label{sec: Architecture}

We aim to keep the model architecture as close to the plain ViT architecture~\cite{dosovitskiy_image_2020} as possible while introducing only the minimal, principled adaptations required for volumetric CT. Our resulting model, SPECTRE, follows a two-stage transformer design consisting of a local ViT~(ViT$_\ell$)  and global ViT~(ViT$_g$) components. ViT$_\ell$ encodes fine-grained, geometry-aware representations from local 3D regions, while ViT$_g$ aggregates these region-level tokens to capture scan-level semantics and long-range dependencies. This hierarchical structure (1)~preserves the simplicity and transferability of the ViT backbone, (2)~enables explicit, low-cost pathways for both localized reasoning and global context modeling, and (3)~retains full compatibility with existing adapter and decoding architectures. SPECTRE benefits from an ad-hoc 3D tokenization process along with custom positional encodings and attention mechanism that tailor the ViTs to the specific requirements of CT imaging. 

\subsection{Minimal 3D Tokenization}\label{sec: Architecture / Tokenization}
Given a CT volume $X \in \mathbb{R}^{H \times W \times D}$, the first stage partitions $X$ into non-overlapping 3D patches $P_i(X)$ and applies a linear projection to obtain token embeddings $T^{(0)}$:
\begin{equation}
    T^{(0)} = \big[\,x_1, \ldots , x_N\,\big]^\top, 
    \qquad x_i = E\big(P_i(X)\big) \in \mathbb{R}^d,
\end{equation}
where $E(\cdot)$ is a linear mapping of the flattened patch. In all experiments, patches are chosen with spatial size~$H_p \times W_p \times D_p = 16 \times 16 \times 8$~voxels. The patch depth~$D_p$ is set to half the in-plane patch size, as voxel spacings in the slice direction are typically about twice as large as those in the axial plane. The embedding dimension is set to $d=1080$, providing a balance between efficiency and resolution. This design corresponds to an effective compression factor of~$\tfrac{2048}{1080} \approx 1.8\overline{962}$.

For a typical volume crop of~$H, W=128$, $D=64$~voxels, representative for many downstream applications (\eg,~lesion segmentation), the above yields 512~tokens, equal to the number of patches obtained from a 2D image of size~$256\times256$ using a patch size of~$16\times16$.

\subsection{Local and Scan-Level Attention}\label{sec: Architecture / Attention}
ViT$_\ell$ implements standard Transformer layers but with attention restricted to local windows. We partition the token grid of a full CT scan into $G$ windows (corresponds to 3D crop of the CT volume), each consisting of $m$ tokens. Each local window is prepended with a learnable $\mathrm{[cls]}$ token~$c_w \in \mathbb{R}^d$ whose final state encodes a compact summary of window-level context. For a window token matrix $T_w \in \mathbb{R}^{(1 + m) \times d}$, we compute Multi-Head Self-Attention~(MHSA) as usual:
\begin{equation}
    \mathrm{Attn}(T_w) = \mathrm{softmax}\!\Big(\frac{QK^\top}{\sqrt{d_k}}\Big) V,
\end{equation}
\begin{equation}
    \qquad Q=T_w W_Q,\; K=T_w W_K,\; V=T_w W_V.
\end{equation}
The per-layer cost across the whole scan is 
\begin{equation}
    \mathrm{Cost}_{\mathrm{global}} = G \cdot \mathcal{O}(m^2 d), 
\end{equation}
which is linear in $G$ for fixed $m$.
After processing by ViT$_\ell$, each window $w$ produces a set of patch tokens~${T^{(\ell)}_{w, i}}_{i=2}^m$ together with a window-level $[\mathrm{cls}]$ token~$T^{(\ell)}_{w, 1}=c_w$. We obtain a compact representation suitable for scan-level aggregation as 
\begin{equation}
    \bar{t}_w = \frac{1}{m - 1} \sum_{i=2}^{m} T^{(\ell)}_{w,i} \in \mathbb{R}^d.
\end{equation}
Then we concatenate $\bar{t}_w$ with the $[\mathrm{cls}]$ token to form a single representation for each window:
\begin{equation}
\label{eq:per_window_concat}
    u_w = \big[\,c_w \,\|\, \bar{t}_w\,\big] \in \mathbb{R}^{2d}, \qquad w=1,\dots,G.
\end{equation}
This design preserves global contextual information from $c_w$ and reduces the token count by summarizing patch-level detail into a single descriptor, thereby controlling memory usage in subsequent stages. The per-window vectors are stacked into a matrix
\begin{equation}
    U = \big[\, u_1, \dots, u_G \,\big]^\top \in \mathbb{R}^{G \times 2d}.
\end{equation}
Before entering the global encoder, the sequence is linearly projected back to dimension $d$ to obtain $\tilde{U}$. A learnable scan-level $[\mathrm{cls}]$ token $c_g \in \mathbb{R}^{d}$ is prepended to $\tilde{U}$ to form the input to the global encoder:
\begin{equation}
    Z = \begin{bmatrix} c_g \\ \tilde{U} \end{bmatrix} \in \mathbb{R}^{(G+1)\times d}.
\end{equation}
The global encoder ViT$_g$ computes full MHSA over $\mathcal{T}_g$:
\begin{equation}
    T^{(g)} = \text{ViT}_g(Z) \in \mathbb{R}^{(G+1)\times d}.
\end{equation}
Because $G \ll m$, attention in $\mathrm{ViT}_g$ remains computationally efficient while aggregating scan-level semantics. 

\subsection{3D Rotary Positional Encoding}\label{sec: Architecture / RoPE}
The proposed architecture employs 3D rotary positional embeddings (RoPE)~\cite{su_roformer_2024}. RoPE injects continuous axial coordinates by rotating query and key vectors rather than by adding learned vectors, which avoids storing large interpolated fields while preserving relative-position information across resolutions. Each attention head has dimension $d_k$ (so $d_k=d/H_s$ or $d_k=d/H_g$ depending on encoder), and by setting
\begin{equation}
    d_k \equiv 0 \pmod{6},
\end{equation}
SPECTRE allocates $L=d_k/6$ frequency slots per axis.

To further improve robustness to resolution and scale changes, RoPE-box jittering is implemented as in DINOv3~\cite{simeoni_dinov3_2025}, applying only a global rescaling with $s \sim \mathcal{U}(0.5,2.0)$ to the normalized coordinates $r_i\in[-1,\ 1]^3$ obtaining $\tilde{r_i}$. With shared frequency periods $p\in\mathbb{R}^{L}$, axis angles are
\begin{equation}
\theta_i^{(a)} = 2\pi,\tilde r_i^{(a)}/p, \quad (a \in {h,w,d}),
\end{equation}
and we define
\begin{equation}
\cos_i=\cos(\Theta_i),\qquad \sin_i=\sin(\Theta_i),
\end{equation}
where $\Theta_i$ is the concatenated per-axis angle vector.

RoPE is then applied to query and key projection heads  via the rotate/merge matrix~$\mathcal{R}$:
\begin{equation}
Q'=\mathcal{R}(Q;\cos_i,\sin_i),\qquad K'=\mathcal{R}(K;\cos_i,\sin_i).
\end{equation}
Using RoPE in both ViT$_\ell$ and ViT$_g$ ensures robustness to local window size, resolution, and varying numbers of windows per scan.
\section{DINO-Driven Vision-Language Pretraining}\label{sec: Pretraining}

We divide pretraining into two complementary stages. The first stage optimizes a self-supervised learning~(SSL) objective that encourages the model to capture fine-grained local visual features from CT volumes. The second stage aligns image-text pairs under weak supervision, extracting semantically meaningful cues from medical reports as well as global scan-level features. Together, these stages enable both detailed spatial understanding and semantic alignment with clinical knowledge. Details about the data pipelines, data preprocessing, hyperparameters, and pretraining hardware can be found in the \emph{Supplementary Material}.

\subsection{Self-Supervised Local Representation Learning}\label{sec: Pretraining / SSL}
The first pretraining stage employs an adapted version of the DINOv3 framework~\cite{simeoni_dinov3_2025}. The framework leverages a student–teacher pair of ViTs built on the ViT$_\ell$ backbone, both of which produce a $\mathrm{[cls]}$ token of embedding size $d=1080$. A three-layer projection head is attached to the backbone to produce the high-dimensional prototypes used for distillation. The student is a masked ViT~\cite{he_masked_2022} where a subset of input patch tokens is replaced by a learnable mask token before encoding. The teacher is an exponential moving average of the student and provides stable soft targets for distillation.

Input views are created by a multi-crop strategy by sampling two global views and eight local views from each input CT volume. Global views are sampled with independent scale ratios on each axis drawn as $r_g \sim \mathcal{U}(0.5,1.0)$ and rescaled to 0.5 of the original volume. Local crops are sampled with ratios $r_\ell \sim \mathcal{U}(0.1875,0.5)$ and rescaled to 0.1875 of the original volume. In addition to random resized cropping, augmentations applied to all views include: (1)~random flipping along each anatomical axis with probability $p=0.5$, (2)~Gaussian sharpening or Gaussian smoothing with $p=0.25$ (mutually exclusive), (3)~gamma intensity transforms with $p=0.25$, (4)~additive Gaussian noise with $p=0.25$, and (5)~random intensity rescaling with $p=0.5$. The rescaling window is sampled by choosing
\begin{equation}
    w_{low} \sim \mathcal{U}(-1000, \, -200), \qquad w_{high} \sim \mathcal{U}(+200, \, +1000),
\end{equation}
and voxel intensities are linearly rescaled to unity range within that window.

Similar to the DINOv2~\cite{oquab_dinov2_2023} and DINOv3~\cite{simeoni_dinov3_2025} implementations, we jointly optimize the DINO~\cite{caron_emerging_2021}, iBOT~\cite{zhou_image_2021} and KoLeo~\cite{sablayrolles_spreading_2018} training objectives with relative weights of~$1:1:0.1$. In contrast to DINOv3, we deliberately omit the Gram loss term (primarily relevant in the optimization of dense features when scaling models to billions of parameters) due to the relatively narrow operational spectrum of the CT domain. The DINO term enforces global consistency between teacher and student across scales. With $\mathcal{G}$ denoting the set of global crops and $\mathcal{H}$ the set of local crops, the DINO loss is computed as the cross-entropy between the teacher’s global soft targets and the student’s predictions on all other views:
\begin{equation}
    \mathcal{L}_{\mathrm{DINO}} = \frac{1}{D}
\sum_{g \in \mathcal{G}}\;
\sum_{\substack{v \in \mathcal{G}\cup\mathcal{H} \\ v \neq g}}\;
\sum_{k=1}^{K}\;
\bigl(-\, q_t^{(k)}(g)\,\log p_s^{(k)}(v)\bigr),
\end{equation}
where $q_t(g)$ is the teacher softmax on the DINO head for global crop $g$, $p_s(v)$ is the student softmax on crop $v$, and $D=\,|\mathcal{G}| \,\big(|\mathcal{G}|+|\mathcal{H}|-1\big)$ the number of combinations.

The iBOT term performs masked patch self-distillation. Let $\mathcal{M}$ denote the set of masked patch positions and let $C$ denote the token vocabulary size for the token-wise head. The iBOT loss is the average token-level cross-entropy over masked patches:
\begin{equation}
    \mathcal{L}_{\mathrm{iBOT}} = \frac{1}{|\mathcal{M}|}\sum_{m\in\mathcal{M}}\sum_{c=1}^{C}
\bigl(-\, q_t^{(c)}(m)\,\log p_s^{(c)}(m)\bigr),
\end{equation}
where $q_t^{(c)}(m)$ are the teacher’s soft targets for token class $c$ at patch $m$ and $p_s^{(c)}(m)$ are the student’s predicted token probabilities at the same position. In this setup, the teacher produces token-level targets from global views in which a fraction of tokens are replaced by a mask token, and the student is trained to predict these targets at the masked positions. We apply masking to 50\% of the global views within each batch and randomly mask a proportion $\rho \sim \mathcal{U}(0.2, 0.7)$ of tokens. This range is higher than the~$\mathcal{U}(0.1,0.5)$ used in DINOv3 and the original iBOT implementation, reflecting the reduced difficulty of the task in 3D, where each token has a larger number of spatial neighbors. We use separate projection heads for the DINO and iBOT objectives and set the number of output prototypes~$K=C=65,536$.

Finally, KoLeo provides a regularization that encourages a uniform spread of the embeddings to prevent collapse and promote effective use of the representation space. For a batch of L2-normalized embeddings $z_i$, the loss identifies, for each embedding, its nearest neighbor $z_{i}^{\mathrm{NN}}$ based on cosine similarity and penalizes small distances:
\begin{equation}
    \mathcal{L}_{\mathrm{KoLeo}} = -\frac{1}{N} \sum_{i=1}^{N} \log \big( \| z_i - z_{i}^{\mathrm{NN}} \|_p + \varepsilon \big),
\end{equation}
where $\|\cdot\|_p$ denotes the $p$-norm ($p=2$) and $\varepsilon$ is a small constant for numerical stability. The nearest neighbor $z_{i}^{\mathrm{NN}}$ is the embedding in the batch with a maximal cosine similarity to $z_i$ that excludes itself.

\subsection{Global Clinical Context Alignment}\label{sec: Pretraining / VLA}
Following self-supervised pretraining of ViT$_\ell$, the full model is aligned with free-text clinical reports using the SigLIP objective~\cite{zhai_sigmoid_2023}. 

Each preprocessed scan is partitioned into $G=36$ windows of size $128 \times 128 \times 64$~voxels, and encoded by the full model (ViT$_\ell$ and ViT$_g$) into a feature vector of embedding size~$d=1080$. Radiology text is encoded with the Qwen3-0.6B Embedding model~\cite{zhang_qwen3_2025} augmented with low-rank adapters~(LoRA; rank~$r=16$, $\alpha=64$), while image and text embeddings are projected into a shared 512-dimensional space using three-layer projection heads. The embeddings are $L_2$-normalized before similarity computation. To allow efficient computation of all image–text pair similarities across a compute cluster, text embeddings are shuffled across devices while image embeddings remain local, allowing the use of a large set of negatives without excessive communication.

The SigLIP loss replaces the softmax-based InfoNCE~\cite{oord_representation_2019} used in conventional methods such as CLIP~\cite{radford_learning_2021} with binary cross-entropy terms based on symmetric sigmoids. The sigmoid-based terms better accommodate the inherently noisy and many-to-many nature of vision-text pairs in clinical datasets, where a single scan may correspond to multiple textual descriptions of varying granularity. Furthermore, losses based on sigmoid have shown to be less sensitive to small batch sizes compared to their softmax-based counterparts~\cite{zhai_sigmoid_2023}.

Let $v_i$ and $t_i$ denote the normalized image and text embeddings for the $i$-th paired sample, and let $\tau > 0$ be a temperature. Define the scaled cosine similarity
\begin{equation}
    \operatorname{sim}(v,t) \;=\; \frac{\langle v, t\rangle}{\tau}.
\end{equation}
The directional SigLIP loss from images to text is written as
\begin{equation}
    \begin{aligned}
        \mathcal{L}_{v\!\to\!t}
        = -\frac{1}{N}\sum_{i=1}^{N} \Bigg[
            &\log\sigma\big(\operatorname{sim}(\tilde v_i,\tilde t_i)\big) \\
            &+ \frac{1}{N-1}\sum_{j\neq i}
              \log\!\Big(1-\sigma\big(\operatorname{sim}(\tilde v_i,\tilde t_j)\big)\Big)
        \Bigg].
    \end{aligned}
\end{equation}
with $\sigma(x)=(1+e^{-x})^{-1}$. The total SigLIP loss averages in both directions as
\begin{equation}
    \mathcal{L}_{\mathrm{SigLIP}} = \tfrac{1}{2}\big(\mathcal{L}_{v\!\to\!t} + \mathcal{L}_{t\!\to\!v}\big),
\end{equation}
where $\mathcal{L}_{t\!\to\!v}$ is defined analogously by swapping image and text embeddings. This symmetric formulation ensures reciprocal alignment, while retaining the robustness advantages of the sigmoid-based objective.

\section{Experiments}\label{sec: Experiments}

\subsection{Cancer Image Biomarker Prediction}\label{subsec:cancer_biomarker}
To assess the discriminative power and generalizability of volumetric representations, we adopt the standardized evaluation protocol of \citet{pai_foundation_2025} and apply it uniformly across all CT foundation models under comparison. For each model, frozen encoder embeddings are extracted and used to train a k-nearest neighbor~(kNN) classifier without finetuning, enabling a controlled comparison of representation quality independent of task-specific optimization.

We assess performance on six datasets spanning diagnostic and prognostic objectives: malignancy prediction on LUNA16~\cite{setio_validation_2017} and DLCS~\cite{wang_duke_2025}, and two-year survival prediction on NSCLC-Radiomics~\cite{aerts_decoding_2014}, NSCLC-Radiogenomics~\cite{bakr_radiogenomic_2018}, C4KC-KiTS~\cite{heller_state_2021}, and Colorectal-Liver-Metastases~\cite{simpson_preoperative_2024}. Each dataset provides patient-level tumor biomarkers derived from volumetric CT scan crops, allowing consistent evaluation across diverse clinical endpoints.

All models are trained and evaluated under identical protocols to ensure fair comparison with prior benchmarks. Quantitative results are summarized in \cref{fig:radar_plot_foundation_models}. Additional implementation details, as well as further quantitative and qualitative analyses are provided in the \emph{Supplementary Material}.

\subsection{Semantic Segmentation}\label{subsec:segmentation}
To assess whether the representations learned by our volumetric CT transformer transfer to dense prediction, we conduct a series of controlled semantic segmentation experiments on established abdominal and renal CT benchmarks. Our goal here is not to build the strongest task-specific segmentation model -- with heavy, decoder-centric engineering -- but to isolate encoder quality under realistic 3D conditions. We therefore follow the Encoder-only Mask Transformer~(EoMT)~\cite{kerssies_your_2025} approach and extend it to the volumetric case: the pretrained SPECTRE encoder produces a set of volumetric tokens, and we instantiate a fixed set of learnable query tokens, one per semantic class in the target dataset. This removes the need for instance-level matching (no Hungarian, no dynamic query allocation) and allows us to supervise the model with standard voxel-wise losses (CE/Dice) on class masks. Importantly, this approach does predict masks at $\frac{1}{4}$ of the input resolution. Specifically, given an input of size $H \times W \times D$, the model predicts masks on a feature grid of size $\frac{H}{4} \times \frac{W}{4} \times \frac{D}{4}$ to keep computation feasible, and the features are subsequently upsampled to $H \times W \times D$ using simple trilinear interpolation for evaluation. We integrate the model as a drop-in replacement in nnU-Net to ensure comparability with widely accepted 3D medical segmentation practice, keeping the rest of the pipeline predominantly the same so that any gain can be attributed to representation quality rather than to the data augmentation, optimization strategy or model decoder. We call this adaptation to semantic segmentation with the EoMT model, Semantic Encoder only Masked Transformer~(SEoMT). An in-depth discussion on the implementation details and a visual representation of the segmentation architecture can be found in the \emph{Supplementary Material}.

We benchmark on a CT-heavy suite that includes KiTS23~\cite{heller_state_2021} (kidney + tumor + cyst), LiTS~\cite{bilic_liver_2023} (liver + lesion), and WORD~\cite{luo_word_2022} (abdominal organs). These datasets jointly probe large, high-SNR organs with stable shape priors, and small, low-contrast tumor targets that are disproportionately sensitive to attention placement. We tune the model parameters on Fold~0 of the KiTS23 dataset during development, and thus exclude that fold from the final result. For all other experiments, across all data folds and datasets, we use \textit{exactly} the same model hyperperameters to ensure a fair comparison with prior work. The quantitative results of our experiments are reported in \cref{tab:segmentation_results}. The \emph{Supplementary Material} presents quantitative results on additional datasets, along with comparisons to prior CT foundation models. It also includes qualitative results across different datasets.
\begin{table}[]
    \centering
    \caption{Segmentation test results over 4 datasets. Average Dice of all five folds of the test datasets. Kidney tumor segmentation on KiTS23 dataset uses 4 folds.}
    \resizebox{0.90\columnwidth}{!}{%
    \begin{tabular}{l|l|rrr}
        \toprule
          & \multirow{2}{*}{\emph{Method}}  & \multicolumn{3}{c}{\emph{Dice (\%) $\uparrow$}}    \\
          &    & \multicolumn{1}{c}{KiTS23} & \multicolumn{1}{c}{LiTS}  & \multicolumn{1}{c}{WORD} \\
        \midrule
        \multirow{2}{*}{\rotatebox[origin=c]{90}{Conv.}} 
        & nnU-Net~\cite{isensee_nnu-net_2021}      & 85.99 & 79.29 & 83.11   \\
        & nnU-Net ResEnc L~\cite{linguraru_nnu-net_2024} & \textbf{88.06} & \textbf{81.20} &  \textbf{85.79} \\
        \midrule
        \multirow{8}{*}{\rotatebox[origin=c]{90}{Transformers}}
        & CoTr~\cite{de_bruijne_cotr_2021}       & 84.63 & 78.44 & 83.11   \\
        & nnFormer~\cite{zhou_nnformer_2023}     & 75.72 & 77.02 &  82.53   \\
        & SwinUNETRv2~\cite{greenspan_swinunetr-v2_2023}   & 75.07 & 73.27 &  78.99     \\
        & UNETR~\cite{hatamizadeh_unetr_2022}         & 76.33 & 70.91 & 70.87 \\
        & WaveFormer~\cite{gee_waveformer_2026}           & 80.61 &  -  & - \\
        & Primus-M~\cite{wald_primus_2025}                & 86.13 & 79.52 & 83.19  \\
        & Primus-L~\cite{wald_primus_2025}                 & 86.04 & 79.90 &  82.99  \\
        \midrule
        & SPECTRE (ours)          & \underline{86.64} &  \underline{80.14}  &   \underline{83.31} \\
        \bottomrule
    \end{tabular}}
    \label{tab:segmentation_results}
\end{table}

\subsection{Zero-Shot Text-to-Image Retrieval}
To evaluate the cross-modal alignment capabilities of our full-scan representations, we perform zero-shot text-to-image retrieval experiments on two large-scale radiology datasets: CT-RATE~\cite{hamamci_foundation_2024} and Merlin~\cite{blankemeier_merlin_2024}. We assess how well the learned embeddings capture clinically meaningful relationships between volumetric CT data and free-text radiology reports, without any task-specific fine-tuning.

Following standard practice, both image and text inputs are projected into a shared latent space, and retrieval is performed by computing cosine similarity between modality embeddings. For CT-RATE, we use the validation split and evaluate recall-based retrieval metrics at various thresholds (Recall@K). The Merlin benchmark further enables analysis of report section granularity, providing Findings-to-Image and Impressions-to-Image retrieval tasks on its provided test split. We additionally evaluate a combined setting where both sections are jointly encoded to form a unified query representation.

All experiments are conducted using the pretrained SPECTRE encoder, Qwen3-0.6B Embedding model with LoRA adapters, and SigLIP projection heads without additional adaptation. Baselines include CT-CLIP~\cite{hamamci_foundation_2024} and the Merlin foundation model~\cite{blankemeier_merlin_2024}, as well as OpenCLIP~\cite{cherti_reproducible_2023} and BioMedCLIP~\cite{zhang_biomedclip_2025}. Quantitative results are summarized in \cref{tab:full report retrieval} and \cref{tab: merlin retrieval}. Several ablation studies on the effects of report noise, report length, CT voxel spacings, as well as results from other SigLIP-based models and UMAP visualizations of the joint embedding space of SPECTRE are provided in the \emph{Supplementary Material}.
\begin{table}[]
    \centering
    \caption{Text-to-image retrieval performance on full radiology reports of the validation set of CT-RATE~\cite{hamamci_foundation_2024} (N=1,564), including both \emph{Findings} and \emph{Impressions} sections.}
    \resizebox{0.75\columnwidth}{!}{%
    \begin{tabular}{l|rrrr}
        \toprule
        \multirow{2}{*}{\emph{Method}}       & \multicolumn{4}{c}{\emph{Recall@K (\%) $\uparrow$}} \\
        & \multicolumn{1}{c}{K=5}   & \multicolumn{1}{c}{K=10}  & \multicolumn{1}{c}{K=50} & \multicolumn{1}{c}{K=100} \\
        \midrule
        CT-CLIP~\cite{hamamci_foundation_2024}  & 2.9   & 5.0   & 18.0   & 28.8 \\
        SPECTRE (ours)   & \textbf{17.5}  &  \textbf{25.5}  & \textbf{48.9}   & \textbf{59.9} \\
        \midrule
        \emph{Random chance}       & 0.3   & 0.6   & 3.2   & 6.4 \\
        \bottomrule
    \end{tabular}}
    \label{tab:full report retrieval}
\end{table}
\begin{table}[]
    \centering
    \caption{Text-to-image retrieval performance on the test set of Merlin~\cite{blankemeier_merlin_2024}, separated for the different \emph{Findings} and \emph{Impressions} sections, and for the \emph{Full Report (=FR)}. Recall is calculated for different sizes of non-overlapping data pools~N.}
    \resizebox{0.90\columnwidth}{!}{%
    \begin{tabular}{l|l|rrr|rrr}
        \toprule
         & \multirow{3}{*}{\emph{Method}} & \multicolumn{3}{c|}{\emph{Recall@1 (\%) $\uparrow$}} & \multicolumn{3}{c}{\emph{Recall@8 (\%) $\uparrow$}} \\
         & & \multicolumn{3}{c|}{N=}    &   \multicolumn{3}{c}{N=}  \\
         & & \multicolumn{1}{c}{32} & \multicolumn{1}{c}{64} & \multicolumn{1}{c|}{128} & \multicolumn{1}{c}{32} & \multicolumn{1}{c}{64} & \multicolumn{1}{c}{128} \\
        \midrule
        \multirow{4}{*}{\rotatebox{90}{\emph{Findings}}} & OpenCLIP~\cite{cherti_reproducible_2023} & 3.3 & 1.7 & 0.9 & 25.0 & 12.5 & 6.1 \\
         & BioMedCLIP~\cite{zhang_biomedclip_2025} & 4.4 & 2.1 & 1.2 & 30.6 & 15.6 & 7.9 \\
         & Merlin~\cite{blankemeier_merlin_2024} & \textbf{77.6} & \textbf{68.7} & \textbf{59.4} & \textbf{98.8} & \textbf{96.5} & \textbf{92.0} \\
         & SPECTRE (ours) & \underline{55.5} & \underline{43.8} & \underline{33.0} & \underline{93.5} & \underline{86.3} & \underline{76.1} \\
        \midrule
        \multirow{4}{*}{\rotatebox{90}{\emph{Impressions}}} & OpenCLIP~\cite{cherti_reproducible_2023} & 3.2 & 1.7 & 0.8 & 25.2 & 12.6 & 6.4 \\
         & BioMedCLIP~\cite{zhang_biomedclip_2025} & 4.6 & 2.4 & 1.2 & 32.2 & 16.9 & 8.1 \\
         & Merlin~\cite{blankemeier_merlin_2024} & \underline{38.4} & \underline{27.7} & \underline{19.4} & \textbf{85.4} & \underline{70.6} & \underline{56.4} \\
         & SPECTRE (ours) & \textbf{43.2} & \textbf{32.9} & \textbf{24.0} & \underline{83.7} & \textbf{72.8} & \textbf{61.5} \\
         \midrule
         \rotatebox{90}{\emph{FR}} & SPECTRE (ours) & 66.4 & 55.7 & 44.8 & 96.4 & 91.6 & 84.3 \\
         \midrule
         & \emph{Random chance} & 3.1 & 1.6 & 0.8 & 25.0 & 12.5 & 6.3 \\
        \bottomrule
    \end{tabular}}
    \label{tab: merlin retrieval}
\end{table}

\section{Discussion}\label{sec: Discussion}
This research scales the pretraining of CT transformers by combining three key elements: a large collection of CT scans paired with radiology reports, transformer architectures specifically optimized for volumetric CT data, and state-of-the-art pretraining methods adapted from recent SSL and VLA advances.

The resulting foundation model, SPECTRE, delivers consistent improvements across diverse tasks. On tumor biomarker classification, our model achieves the best overall performance among prior CT foundation models, outperforming competitors on 4/6 tasks. This demonstrates that large-scale paired pretraining yields representations that capture both structural and clinically meaningful information, enabling better discrimination of subtle biomarker-related patterns. For segmentation, our encoder-only framework surpasses all other domain-specific transformer-based models and performs competitively with nnU-Net, despite lacking any decoder-heavy design. Because the segmentation head simply interpolates the embedding vectors, the resulting predictions are remarkably smooth and anatomically coherent. However, this design can miss high resolution details when necessary - an area for improvement.

Beyond these supervised evaluations, SPECTRE exhibits strong cross-modal alignment. In full-scan text-to-image retrieval, it substantially outperforms CT-RATE, indicating robust visual–textual grounding. When analyzing retrieval from specific report sections, MERLIN performs best on the structured Findings text but struggles with the more interpretive Impressions section. In contrast, our model achieves the best results on Impressions-to-image retrieval and improves further when combining sections (Findings + Impressions), suggesting stronger integration across textual domains and less sensitivity to report structure. We attribute this to the language rewrites and text augmentations used during SigLIP-style pretraining, which enhance robustness to stylistic and structural variability.

Despite these advances, several limitations remain. Our pretraining corpus, though large, was skewed toward thoracic imaging, which may partly explain the stronger performance on lung-related tasks compared to abdominal ones. The reliance on clinical reports introduces inherent noise and bias, as descriptions vary in completeness and terminology across institutions. Furthermore, while the encoder-only segmentation approach offers elegance and computational efficiency, its smooth outputs can sometimes obscure small or faint lesions. Finally, even though training a large-scale foundation model requires substantial computational resources, sharing the trained model publicly enables broad reuse and mitigates the need for repeated training efforts. By releasing SPECTRE as an open, general-purpose 3D CT foundation model, we lower the barrier for institutions worldwide, enabling more accurate, data-efficient, and locally adapted medical imaging tools that can improve global health equity.

\section{Conclusion}\label{sec: Conclusion}
We presented SPECTRE, a fully transformer-based foundation model for volumetric CT and radiological reports understanding. The geometry-consistent, two-stage 3D Vision Transformer trained with large-scale self-supervision and subsequent report-level vision–language alignment with our text encoder, constitutes a single, task-agnostic volumetric backbone that performs well across heterogeneous medical imaging objectives. Without any decoder-heavy redesign, the pretrained 3D encoder both maintains SOTA-level frozen transfer on 4/6 CT biomarker benchmarks and, in its encoder-only SEoMT form, reaches competitive performance on dense semantic segmentation tasks. At the same time, it preserves spatial discriminability under vision–language alignment, delivering strong CT–report retrieval and understanding. This establishes a scalable path toward general-purpose 3D medical foundation models.
\section*{Acknowledgments}
The authors acknowledge the Supercomputing Center of the Eindhoven University of Technology (\url{www.supercomputing.tue.nl/}) for providing access to and assistance with the various computing resources available. We further acknowledge SURF (\url{www.surf.nl}) for their assistance in enabling the use of the Dutch national supercomputer Snellius.
\appendix
\clearpage
\setcounter{page}{1}
\maketitlesupplementary

\section{Pretraining Data and Preprocessing}
\label{app: datasets}

\begin{table}[]
    \centering
    \small
    \caption{Overview of the datasets used for pretraining, summarizing anatomical coverage, the number of CT reconstructions remaining after all exclusions, and whether each dataset is used for self-supervised learning~(SSL), vision–language alignment~(VLA), or both.}
    \resizebox{0.9\columnwidth}{!}{%
    \begin{tabular}{l|lrcc}
        \toprule
        \emph{Source dataset}                               & \emph{Location}       & \emph{No. Rec.} & \emph{SSL} &\emph{VLA} \\
        \midrule
        NLST~\cite{the_nlst_research_team_reduced_2011}    & Chest                 & 132,985               & \cmark & \\
        CT-RATE~\cite{hamamci_foundation_2024}             & Chest                 & 47,149                & \cmark & \cmark \\
        INSPECT~\cite{huang_inspect_2023}                  & Chest                 & 23,226                & \cmark & \cmark  \\
        Merlin~\cite{blankemeier_merlin_2024}              & Abdomen               & 15,314                & \cmark & \cmark \\
        AbdomenAtlas~\cite{li_abdomenatlas_2024}    & Abdomen               & 5,195                 & \cmark &  \\
        AMOS~\cite{ji_amos_2022}                           & Abdomen               & 2,450                 & \cmark &  \\
        PANORAMA~\cite{alves_panorama_2024}                & Abdomen               & 2,238                 & \cmark &  \\
        AbdomenCT-1K~\cite{ma_abdomenct-1k_2022}           & Abdomen               & 1,062                 & \cmark &  \\
        \midrule
        \textbf{Total}                                      &                       & \textbf{229,619}        & &  \\
        \bottomrule
    \end{tabular}}
    \label{tab:datasets}
\end{table}

\subsection{Datasets}
We curated a diverse collection of 3D CT scans from multiple publicly available datasets. Three of these datasets also include accompanying clinical metadata, such as radiology reports and EHR diagnostic codes, and can therefore be used for both SSL and VLA. Imaging data span the thoracic, abdominal, and pelvic regions and comprise a wide range of acquisition settings, including variations in radiation dose and the use of contrast agents. After applying exclusion criteria that are provided below for every dataset, the final set for pretraining comprises 229,619~image series. A summary of the datasets and their characteristics is provided in \cref{tab:datasets}.

We now provide dataset-level summaries, including the filtering criteria and preprocessing steps used to construct the final pretraining corpus.
\begin{itemize}
    \item \textbf{NLST}~(the National Lung Screening Trial)~\cite{the_nlst_research_team_reduced_2011} provides lung cancer screening data collected in the United States between 2002~and~2004. The dataset consists of low-dose helical chest CT scans from 26,254~participants with a two-year follow-up, yielding 73,116~studies. Owing to multiple reconstruction settings, the original release contains 203,099~series. For our purposes, we retain only one series per reconstruction kernel; if multiple series are available for the same kernel, we select the one with the largest number of slices. This yields at most two series per CT study, for a total of 132,985~image series. All relevant DICOM series are converted into 3D volumes in NiFTi format.
    
    \item \textbf{CT-RATE}~\cite{hamamci_foundation_2024} provides paired 3D chest CT volumes and the corresponding radiology reports of 21,304~patients. It comprises 25,692~non-contrast chest CT studies, expanded to 50,188~series through multiple reconstructions. Each study is accompanied by the radiology report, including both \emph{Findings} and \emph{Impressions} sections, as well as multi-abnormality labels and metadata. The cohort is split into 20,000~patients for training and 1,304~patients for validation. To ensure data consistency, we excluded unintended head CT series by generating segmentation masks with the TotalSegmentator model~\cite{wasserthal_totalsegmentator_2023} and removing scans where the proportion of voxels labeled as \emph{brain} or \emph{skull} was an outlier, with outliers defined by Tukey’s rule ($\mathrm{Q3} + 1.5 \times \mathrm{IQR}$) of the distribution of relative brain/skull volume. We also respect the original split and use only the original training set for pretraining of our model, resulting in a total of 47,149~image series for training.
    
    \item \textbf{INSPECT}~\cite{huang_inspect_2023} consists of CT pulmonary angiography~(CTPA) scans paired with radiology reports that include the \emph{Impressions} section. It contains imaging data from 19,402~patients with a total of 23,248~studies. To address data issues, we excluded partially uploaded files, resulting in a final set of 23,226~CTPA studies.  
    
    \item \textbf{Merlin}~\cite{blankemeier_merlin_2024} contains abdominal CT scans acquired at the Stanford Hospital Emergency Department between 2012 and 2018. It includes 25,494~studies from 18,317~patients, each paired with a radiology report comprising sections \emph{Findings} and \emph{Impressions}, as well as associated EHR diagnostic codes. The data is split into training, validation, and test sets, with 15,314~studies in the training set. We use the training split of the dataset in its provided form without applying further filtering or modifications to train our model.
    
    \item \textbf{AbdomenAtlas1.0Mini}~\cite{li_abdomenatlas_2024} is a fully annotated publicly accessible subset of the larger AbdomenAtlas dataset, comprising 5,195~abdominal CT volumes with segmentations at the voxel-level. The annotations cover nine key anatomical structures: spleen, liver, left kidney, right kidney, stomach, gallbladder, pancreas, aorta, and inferior vena cava. The source images are aggregated from multiple existing public datasets, each of which contributes cases with varying imaging protocols, disease states, and anatomical coverage. For pretraining, we use only the raw CT scans without the accompanying segmentation labels.
    
    \item \textbf{AMOS}~\cite{ji_amos_2022} is a multi-center dataset designed for multi-organ abdominal segmentation across diverse clinical scenarios. It comprises 500~CT and 100~magnetic resonance images~(MRI) with voxel-level annotations for 15~abdominal organs, collected from multi-vendor and multi-phase acquisitions spanning a wide range of disease conditions. In addition, AMOS provides 1,900~unlabeled CT and 1,200~unlabeled MRI scans to support semi-supervised and unsupervised learning tasks. After excluding 50~corrupted or incomplete CT files, we retain a total of 2,450~CT scans for pretraining.
    
    \item \textbf{PANORAMA}~\cite{alves_panorama_2024} is a contrast-enhanced abdominal CT dataset designed to benchmark diagnostic performance for pancreatic ductal adenocarcinoma~(PDAC) detection and diagnosis. It includes 2,238~anonymized CT scans acquired at two Dutch medical centers (Radboud University Medical Center and University Medical Center Groningen). The dataset was curated to ensure high-quality imaging and standardized acquisition protocols.
    
    \item \textbf{AbdomenCT-1K}~\cite{ma_abdomenct-1k_2022} is a large and diverse abdominal CT dataset comprising 1,062~scans collected by aggregating multiple public single-organ datasets. It includes both contrast-enhanced and non-contrast studies with voxel-level annotations for four major abdominal organs: liver, kidneys, spleen, and pancreas. For our purposes, we use only the raw CT scans for pretraining.
\end{itemize}

\subsection{Image Processing for Self-Supervised Learning}
For SSL with the adapted DINOv3, all CT series listed in \cref{tab:datasets} are first reoriented to a common anatomical coordinate system~(right-left, anterior-posterior, superior-inferior; RAS) to ensure spatial consistency. The voxel intensities in Hounsfield Units~(HU) are clipped to the range~$[-1000, +1000]$ to remove outliers and normalized to the unit range with 32-bit precision. Each scan is then resampled to a voxel spacing of $0.5 \times 0.5 \times 1.0$~mm using trilinear interpolation and center-cropped to a maximum size of $512 \times 512 \times 384$~voxels, sufficient to cover the FOV of most chest and abdominal scans while minimizing surrounding background. The resulting volumes are converted to 16-bit tensors and stored on disk. During training, these tensors are loaded into memory and a random crop of $256 \times 256 \times 128$~voxels is extracted for each batch element. To further improve I/O efficiency, four random crops are sampled from the same CT scan and treated as separate batch elements.  

\subsection{Data Processing for Vision-Language Alignment}
For VLA with SigLIP, we first select the CT series that contain accompanying text from radiology reports or diagnostic codes (see \cref{tab:datasets}). Following, all series are reoriented to the RAS coordinate system, clipped within the range~$[–1000,+1000]$~HU and normalized to a unity range, followed by resampling to a voxel spacing of $0.75 \times 0.75 \times 1.5$~mm using trilinear interpolation. Again, we write the resulting tensors in 16-bit precision to disk and load accordingly during training.

To enhance textual descriptions, each paragraph in the radiology reports is expanded to multiple paraphrases. First, the reports are divided into two sections, \emph{Findings} and \emph{Impressions}, whenever these sections are provided. For each section, we prompt a large language model~(LLM) to  ``rephrase clearly and concisely \emph{without changing any medical facts},'' and to ``\emph{only} return the revised text'', ensuring clarity and consistency.

This prompt is supported by four examples curated by radiologists~(two chest CT, two abdominal CT), demonstrating accurate rewrites that preserve clinical elements such as laterality, measurements, and negations. Using Google's~\emph{Gemini~2~Flash}, we generate two additional paraphrases per section, resulting in three semantically equivalent versions. During training, a version is randomly sampled, following the LaCLIP single positive strategy~\cite{fan_improving_2023}, to provide clinically accurate supervision of vision-language alignment. 

In addition to text, some reports include structured EHR diagnosis codes. Since LLMs struggle with raw codes (\emph{e.g.}, $\mathrm{J18.9}$), we replace each with its World Health Organization 2025 short description\footnote{\url{https://www.who.int/standards/classifications/classification-of-diseases}}. These are appended as a comma separated list at the end of the report to form a complete text input. The EHR descriptions are not rewritten and are added after LLM rewriting to preserve billing and epidemiological accuracy.

\section{Pretraining \& Architectural Details}
\label{app: training details}
\begin{table}[]
    \centering
    \caption{Training hyperparameters during self-supervised learning~(SSL) and vision–language alignment~(VLA) pretraining stages; LR, LLRD, and WD denote learning rate, linear learning rate decay, and weight decay.}
    \resizebox{0.9\columnwidth}{!}{%
    \begin{tabular}{l|cc}
    \toprule
    \emph{Hyperparameter}                   & \emph{SSL}                                & \emph{VLA}  \\
    \midrule
    Epochs                                      & 150                                       & 100  \\
    Steps                                       & 83,100                                    & 59,000      \\
    Batch size                                  & 1,536                                     &  144 \\
    LR begin                              & $4 \times 10^{-3}$                        & $1 \times 10^{-4}$  \\
    LR end                           & $1 \times 10^{-6}$                        & $1 \times 10^{-6}$  \\
    LR schedule                      & Cosine annealing                          & Cosine annealing  \\
    LR warmup epochs                 & 10                                        & 10  \\
    LLRD                  & 0.9                                       & 0.9  \\
    WD begin                               & 0.04                                      & 0.01  \\
    WD end                            & 0.4                                       & 0.01  \\
    WD schedule                       & Inverse cosine annealing                  & Constant  \\
    \multirow{2}{*}{Optimizer}        & AdamW     & AdamW  \\
    & ($\beta_1=0.9$, $\beta_2=0.999$) & ($\beta_1=0.9$, $\beta_2=0.95$) \\
    \bottomrule
    \end{tabular}}
    \label{tab:train cofig}
\end{table}

\begin{table}[]
    \centering
    \caption{Architectural and model-scale specifications of the local (ViT$_\ell$) and global (ViT$_g$) parts of SPECTRE.}
    \resizebox{0.9\columnwidth}{!}{%
    \begin{tabular}{l|cc}
    \toprule
    \emph{Configuration}                      & \emph{ViT$_{\ell}$}                       & \emph{ViT$_g$}  \\
    \midrule
    Trainable parameters                        & 339M                                      & 58M  \\
    Patch size                                  & $16 \times 16 \times 8$                   & - \\
    Layers                                      & 24                                        & 4  \\
    Embedding dimension                         & 1,080                                     & 1,080  \\
    Attention heads                             & 12                                        & 12  \\
    Position embedding                          & 3D RoPE                                   & 3D RoPE  \\
    LayerScale initialization                   & 0.1                                       & 1.0  \\
    \bottomrule
    \end{tabular}}
    \label{tab:model cofig}
\end{table}

\subsection{Training and Model Configuration}
\cref{tab:train cofig} summarizes the training parameters for both SSL with DINOv3 and VLA with SigLIP. During VLA, the pretrained ViT$_{\ell}$ remains frozen for the first 10~epochs. To improve efficiency, all models are trained with mixed precision~(FP16) and optimized using distributed data parallelism. Data loading is performed via GPU Direct Storage, enabling high-throughput I/O directly to device memory and thereby minimizing bottlenecks in large-scale training. \cref{tab:model cofig} shows the architectural choices of SPECTRE, separated for the local ViT$_\ell$ and global ViT$_g$. Note that during VLA, the LayerScale layers of ViT$_{\ell}$ are initialized with the weights found during SSL.

\subsection{Hardware}
Both pretraining phases of the foundation model are conducted on a cluster of three DGX~B200 systems~(NVIDIA~Corp., CA, USA), totaling 24~Blackwell GPUs with 4.32~TB of combined GPU memory. Each system contains 8~B200 GPUs (1.44~TB per DGX), dual Intel Xeon Platinum~8570 processors (112~cores, 224~threads), and 2.16~TB of system memory, yielding a cumulative 6.48~TB across the cluster. The three DGX systems are interconnected via high-speed InfiniBand, enabling efficient distributed training and data exchange.

\section{Downstream Experiments}\label{app:downstream}
\subsection{Cancer Image Biomarker Prediction}\label{app:downstream/cancer_biomarker}
This section complements the analyses on the \emph{Cancer Image Biomarker Prediction} experiments and provides additional details on the benchmark tasks, datasets, and evaluation setup used in the downstream cancer imaging experiments.

\subsubsection{Foundation Models}
We compare a comprehensive set of eleven publicly available CT foundation models: FMCIB~\cite{pai_foundation_2024}, CT-FM~\cite{pai_vision_2025}, CT-CLIP~\cite{hamamci_foundation_2024}, PASTA~\cite{lei_synthetic_2025}, VISTA3D~\cite{he_vista3d_2025}, VOCO~\cite{wu_voco_2024}, SUPREM~\cite{li_how_2025}, Merlin~\cite{blankemeier_merlin_2024}, MedImageInsight~\cite{codella_medimageinsight_2024}, ModelsGenesis~\cite{zhou_models_2021}, and our proposed SPECTRE. These models collectively represent the current generation of volumetric CT foundation models, spanning both unimodal (image-only) and multimodal (image–text) pretraining paradigms. More about these models can be found in the \emph{Related Works} of the main paper and in the models' respective papers.

\subsubsection{Evaluation Framework}
All models are evaluated within the standardized \emph{TumorImagingBench} reference framework introduced by \citet{pai_foundation_2025}. This framework ensures that embeddings are extracted under consistent preprocessing conditions, reproducing the \emph{intensity normalization}, \emph{crop sizes}, and \emph{voxel spacings} used during each model’s original pretraining. Such harmonized extraction allows for direct comparison of representation quality across models without task-specific retraining. For SPECTRE, we use a default input size of $128 \times 128 \times 64$~voxels with a voxel spacing of $0.5 \times 0.5 \times 1.0$~mm. Since SPECTRE is trained agnostic to input crop size and spacing, we double the field of view for the \emph{NSCLC-Radiogenomics}~\cite{bakr_radiogenomic_2018} and \emph{Colorectal-Liver-Metastases}~\cite{simpson_preoperative_2024} datasets to ensure that all lesions are fully contained within the input volume. All other datasets use the default configuration.

\subsubsection{Tasks \& Datasets}
The TumorImagingBench spans six public datasets covering diagnostic and prognostic tasks in thoracic, renal, and hepatic oncology. The benchmark includes two task types: (1)~lung nodule malignancy classification, and (2)~prediction of two-year survival across multiple tumor sites. A brief overview of the datasets used in our experiments is provided below.
\begin{itemize}
  \item \textbf{LUNA16}~\cite{setio_validation_2017}. A dataset containing 888 CT scans and 1,186~annotated lung nodules. We follow the established subset of 677 nodules enriched for malignancy suspicion. Task: \emph{malignancy classification}.
  \item \textbf{DLCS}~(Duke Lung Cancer Screening)~\cite{wang_duke_2025}. A clinical lung-nodule cohort with 2,487~nodules from 1,613~patients; we adopt the publicly released portion with 1,714 scans and pathology-confirmed malignancy labels. Task: \emph{malignancy classification}.
  \item \textbf{NSCLC-Radiomics}~\cite{aerts_decoding_2014}. CT scans from 421~patients with stage~I–IIIB non-small cell lung cancer~(NSCLC) treated with radiation therapy, including expert Gross Tumor Volume (GTV) segmentations. Task: \emph{two-year survival prediction}.
  \item \textbf{NSCLC-Radiogenomics}~\cite{bakr_radiogenomic_2018}. Surgical NSCLC cohort with preoperative CT/PET imaging; we use 133~cases with curated GTV segmentations. Task: \emph{two-year survival prediction}.
  \item \textbf{C4KC-KiTS}~\cite{heller_state_2021}. Renal tumour cohort from partial or radical nephrectomy patients; after filtering for complete segmentations and follow-up, 134~cases remain. Task: \emph{two-year survival prediction}.
  \item \textbf{Colorectal-Liver-Metastases}~\cite{simpson_preoperative_2024}. Preoperative CT scans from 194~patients undergoing resection of colorectal liver metastases, using the largest lesion per patient. Task: \emph{two-year survival prediction}.
\end{itemize}

\subsubsection{Analysis}
\begin{figure*}
    \centering
    \includegraphics[width=0.9\linewidth]{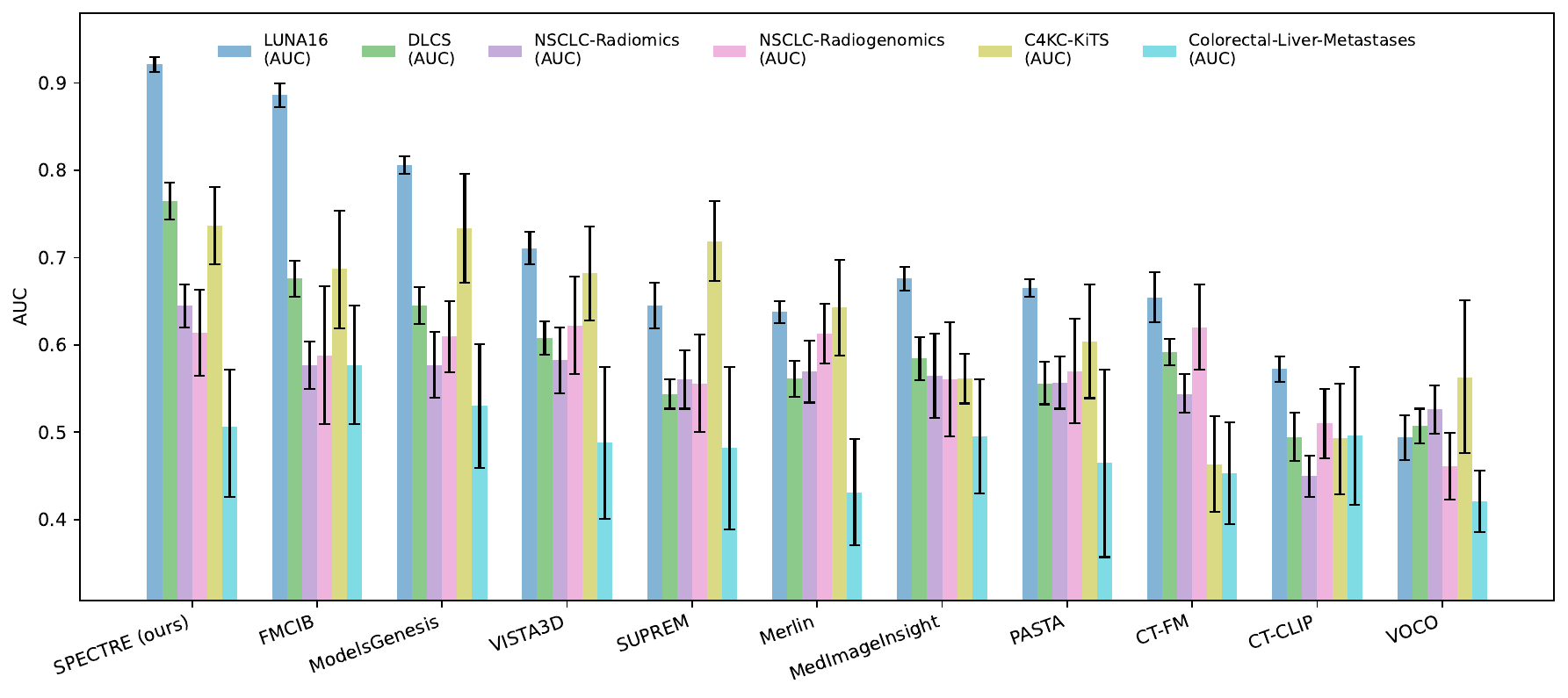}
    \caption{Quantitative comparison of 11~CT~foundation models across six~biomarker classification benchmarks using frozen-embedding kNN classifiers. Bars represent mean performance for each task, with error bars indicating 95\%~confidence intervals across cross-validation folds.}
    \label{fig:bar_auc_foundation_models}
\end{figure*}

Further quantitative analyses on these tasks are provided in \cref{fig:bar_auc_foundation_models}, which reports per-model performance with 95\%~confidence intervals. Notably, for LUNA16, DLCS, and NSCLC-Radiomics, tasks on which our model outperforms all competing approaches, the confidence intervals are narrow, indicating stable performance and low variance across cross-validation folds. In contrast, for NSCLC-Radiogenomics and Colorectal-Liver-metastases, tasks where we do not achieve SOTA performance, all models exhibit large confidence intervals and generally low scores. This is likely due to the smaller dataset sizes, which can introduce quantization noise in the AUC calculation and reflect the inherent difficulty of these tasks.

\begin{figure*}
    \centering
    \resizebox{1.6\columnwidth}{!}{%
    \includegraphics[width=\linewidth]{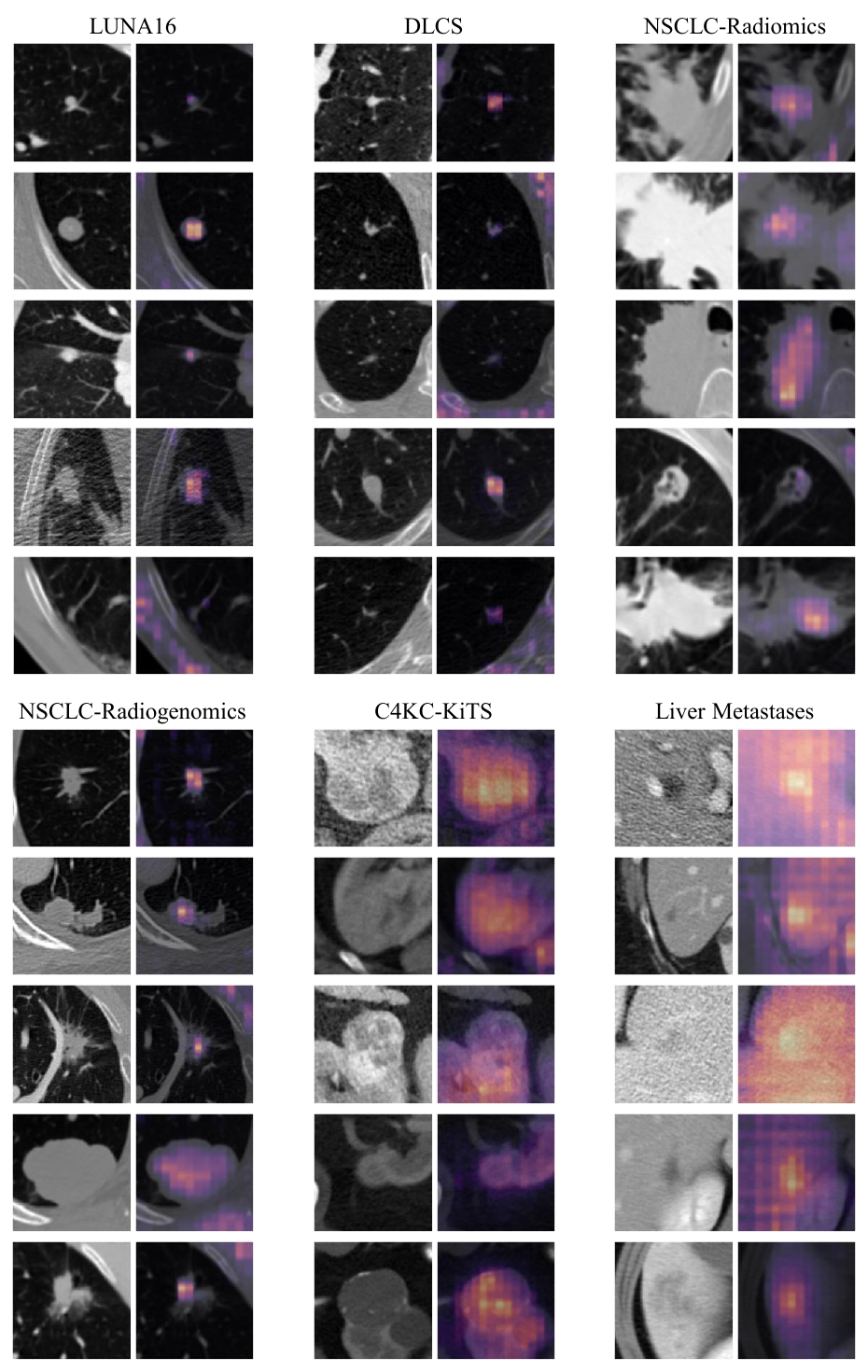}
    }
    \caption{Non-curated saliency maps of SPECTRE on six tumor image biomarker datasets, obtained by occlusion sensitivity.}
    \label{fig:saliency_lung_cancer}
\end{figure*}

Additional qualitative evidence is shown in \cref{fig:saliency_lung_cancer}, which visualizes model explanations using saliency maps on non-curated CT samples. Without any task-specific finetuning, the model already attends to pathologic regions associated with tumor presence, indicating that the learned representations encode clinically relevant spatial features. This behavior supports the effectiveness of our pretraining strategy and echoes findings from earlier foundation-model studies demonstrating that large-scale contrastive or multimodal pretraining facilitates robust zero-shot localization and biomarker-related signal emergence~\cite{pai_foundation_2025}.


\subsection{Semantic Segmentation}\label{appendix:subsec:segmentation}
This section details the full protocol used to evaluate SPECTRE on volumetric \emph{Semantic Segmentation} and makes explicit the detailed experiments that led to the final SEoMT configuration reported in the main paper and the results across the benchmarks obtained.

\subsubsection{Adapting EoMT to 3D Semantic Segmentation} To isolate the segmentation capability of the vision encoder itself, and not any task-specific head, we extend the Encoder-only Mask Transformer~(EoMT) paradigm to volumetric (3D) semantic segmentation. In line to the original 2D EoMT, we remove all task-specific decoders and operate entirely on the encoder token space, also in the 3D case. The model starts from our SPECTRE 3D encoder (ViT$_\ell$), which produces a sequence of anisotropic 3D~tokens (CT patches) using the same tokenizer and 3D~RoPE as in pretraining.

\begin{figure}[h]
    \centering
    \includegraphics[width=0.9\linewidth]{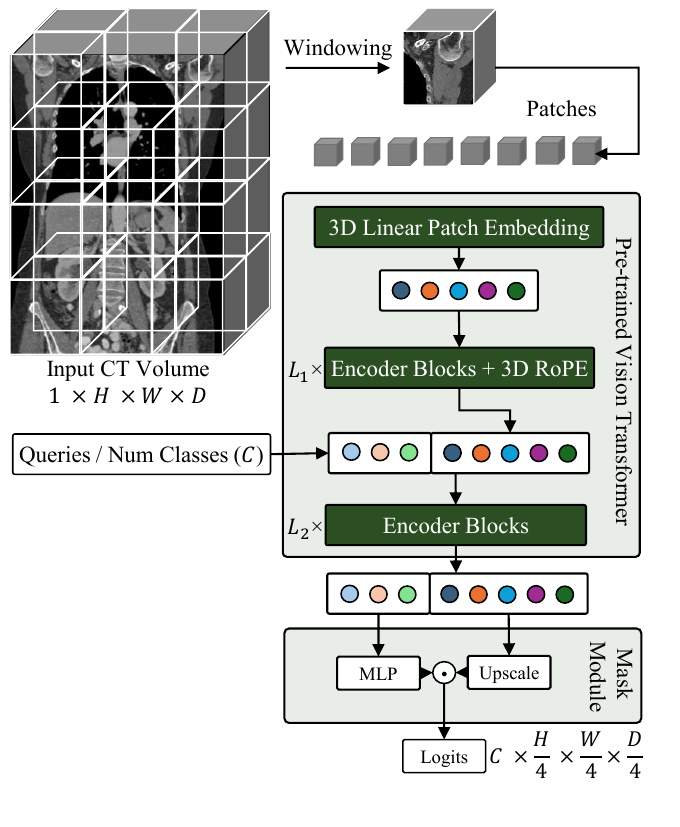}
    \caption{SEoMT architecture, derived from the EoMT. A learnable query for each class C is initialized and concatenated to the patch tokens. The new set of tokens are jointly processed by the last $L_2$ blocks and used to predict logits corresponding to the semantic masks.}
    \label{fig:seomt}
\end{figure}

\subsubsection{Query Design for Semantic Segmentation (3D)} After an initial set of encoder blocks, we append a fixed set of learnable query tokens to this sequence. Because semantic segmentation does not require instance enumeration, we set the number of learnable query tokens equal to the number of semantic classes in the dataset (\eg, 3~for liver, kidney, tumor). The remaining 3D encoder blocks then run joint self-attention over both volume tokens and class queries. This allows the queries to attend to spatial tokens and, symmetrically, lets spatial tokens condition on the queries, so no extra transformer decoder is required. At the output, we obtain (1)~per-class embeddings from the queries and (2)~a dense 3D feature grid from the encoder tokens. We project the 1/4-resolution feature grid to per-voxel class logits and trilinearly upsample it back to the original CT resolution to compute Dice and Cross-Entropy losses. Because the number of classes in medical CT is small and fixed, every query is forced to explain a coherent anatomical or lesion region, which stabilizes training and removes the need for Hungarian matching or instance-slot allocation.

\subsubsection{Integration into nnU-Net}
To position this as a fair encoder-only test, we integrate SPECTRE directly into nnU-Net as a drop-in encoder replacement. Apart from replacing the encoder, no architecture-specific components (multi-scale FPN-style features, convolutions for scale mixing, mask transformer decoders, etc.) are introduced. This ensures the comparison measures ``representation quality of the encoder” -- not engineering around it. We overwrite some of the suggested training plans with new SPECTRE plans. The images are resampled to $0.75\times0.75\times1.5$mm and intensities are rescaled to 0-1 using the 0.5\% and 99.5\% datasets intensity profiles. Additionally we employ the optimizers and learning rate schedulers as suggested in \cite{kerssies_your_2025}, with an AdamW optimizer with a learning rate of $1 \times 10^{-5}$, weight decay $3 \times 10^{-5}$ and gradient clipping of~1.0. Models are trained for 150~epochs with 250~steps per epoch and a batch size of~2, following the noSLL~\cite{wald_openmind_2025} finetuning pipeline. The nnU-Net with SPECTRE integration is publicly available at \url{https://github.com/cviviers/nnUNet}. 

\subsubsection{Datasets}
To avoid unstable conclusions caused by noisy, small, or historically under-annotated radiology datasets, we follow the recommended large-scale segmentation benchmarks by \citet{linguraru_nnu-net_2024}. However, to avoid data contamination, we drop the AMOS dataset as we used it for pretraining. Since our approach focuses purely on CT imaging, we also drop the datasets that contain MRI and add an additional CT dataset. Specifically we consider the datasets as provided in \cref{tab: seg_datasets}.
\begin{table}[]
    \centering
    \caption{Segmentation benchmark datasets. TS = TotalSegmentator.}
    \resizebox{0.9\columnwidth}{!}{%
    \begin{tabular}{lcll}
        \toprule
        \emph{Dataset} & \emph{\# Volumes} & \emph{Classes} & \emph{Description} \\
        \midrule
        KiTS23 & 489 & 3 & kidney / tumor / cyst \\
        LiTS & 131 & 2 & liver + tumor \\
        WORD & 120 & 16 & 16 abdominal organs \\
        TS v1-Full & 1204 & 104 & anatomical structures \\
        TS v201-Full & 1228 & 117 & anatomical structures \\
        TS v201-Merlin & 1228 & 20 & anatomical structures \\
        \bottomrule
    \end{tabular}}
    \label{tab: seg_datasets}
\end{table}

\subsubsection{Evaluation Protocol \& Results}
We adopt the evaluation protocol employed in \citet{wald_primus_2025}. All experiments are conducted within the nnU-Net framework~\cite{isensee_nnu-net_2021}, with the training set randomly divided into 80\%/20\% train/validation splits across 5~folds. After training, the model with the best pseudo Dice is used and validation on the validation set is automatically performed. We directly record the outcome of that result and thus the average of 5-fold cross-validation. For KiTS23, we tuned SPECTRE on \emph{fold-0} during development and exclude that fold from the final reported cross-validation to avoid optimism. All other folds and datasets use exactly the same hyperparameters to make the cross-dataset comparison meaningful.

We compare SPECTRE against various 3D domain-specific segmentation architectures. Specifically, we consider nnU-Net~\cite{isensee_nnu-net_2021} and the updated state-of-the-art ResNet-based nnU-Net ResEnc Large~\cite{linguraru_nnu-net_2024} for comparison with convolutional-based models. Recently many transformer-based models have been developed for segmentation in 3D data. We include CoTr~\cite{de_bruijne_cotr_2021}, nnFormer~\cite{zhou_nnformer_2023}, SwinUNETRv2~\cite{greenspan_swinunetr-v2_2023}, UNETR~\cite{hatamizadeh_unetr_2022}, WaveFormer~\cite{gee_waveformer_2026} and the recent Primus~\cite{wald_primus_2025} model for comparison. The results of these models on KiTS23 and LiTS and WORD are obtained from \citet{wald_primus_2025}. In their experiments, the models were tuned on fold-0 of KiTS23 and LiTS, and thus, the average of the other four folds are used as the baselines. All results are reported in average Dice across all classes. 

During model development, we evaluated the impact of optimization strategies and input crop sizes on downstream segmentation performance after finetuning. The corresponding ablation results are reported in \cref{tab:kits_ablation_segmentation}. Within the nnU-Net framework, Stochastic Gradient Descent~(SGD) constitutes the default optimizer; however, our experiments show that AdamW~\cite{loshchilov_decoupled_2018} consistently outperforms SGD. Additionally, applying Deep Supervision, the default implementation in nnU-Net which computes weighed segmentation losses on intermediate lower resolution layers of the model, further improves training stability and final segmentation performance. We additionally observed that, at smaller crop sizes, models initialized with VLA~(SigLIP) weights outperform counterparts initialized with just SSL~(DINO). Increasing the crop size improves overall performance across initializations, while the performance gap between SSL- and VLA-initialized models narrows, rendering them comparable. Based on these observations, all subsequent experiments employ AdamW with a learning rate of~$1 \times 10^{-5}$, Deep Supervision enabled, and an input crop size of~$320 \times 320 \times 128$.
\begin{table}[]
    \centering
    \small
    \caption{Ablation experiments with the Kidney tumor segmentation on KiTS23~\cite{heller_state_2021} dataset. Results in Dice on fold-0. DNF = ``did not finish".}
    \resizebox{0.9\columnwidth}{!}{%
    \begin{tabular}{l|cc|c}
        \toprule
                                       & \emph{SSL}    & \emph{VLA}  & Dice $\uparrow$  \\
        \midrule
        \multicolumn{2}{l}{\emph{Optimizer} @ $320 \times 320 \times 128$ input crop }\\
        \midrule
        SGD with LR $1 \times 10^{-4}$    & \cmark        & \cmark & DNF \\
        SGD with LR $1 \times 10^{-5}$    & \cmark        & \cmark & 0.763 \\ 
        AdamW  with LR $1 \times 10^{-5}$ & \cmark        & \cmark & 0.868 \\
        \hspace{1em} + Deep Supervision & \cmark        & \cmark & \textbf{0.871} \\
        \midrule
        \emph{Crop Size} & \\
        \midrule
        $128 \times 128 \times 64$                           & \cmark        &           & 0.854 \\
        $128 \times 128 \times 64$                            & \cmark        & \cmark     & 0.862 \\
        $320 \times 320 \times 128$                           & \cmark        &           & \textbf{0.871} \\
        $320 \times 320 \times 128$                            & \cmark        & \cmark     & \textbf{0.871} \\
        \bottomrule
    \end{tabular}}
    \label{tab:kits_ablation_segmentation}
\end{table}
\begin{figure*}
    \centering
    \resizebox{1.6\columnwidth}{!}{%
    \includegraphics[width=\linewidth]{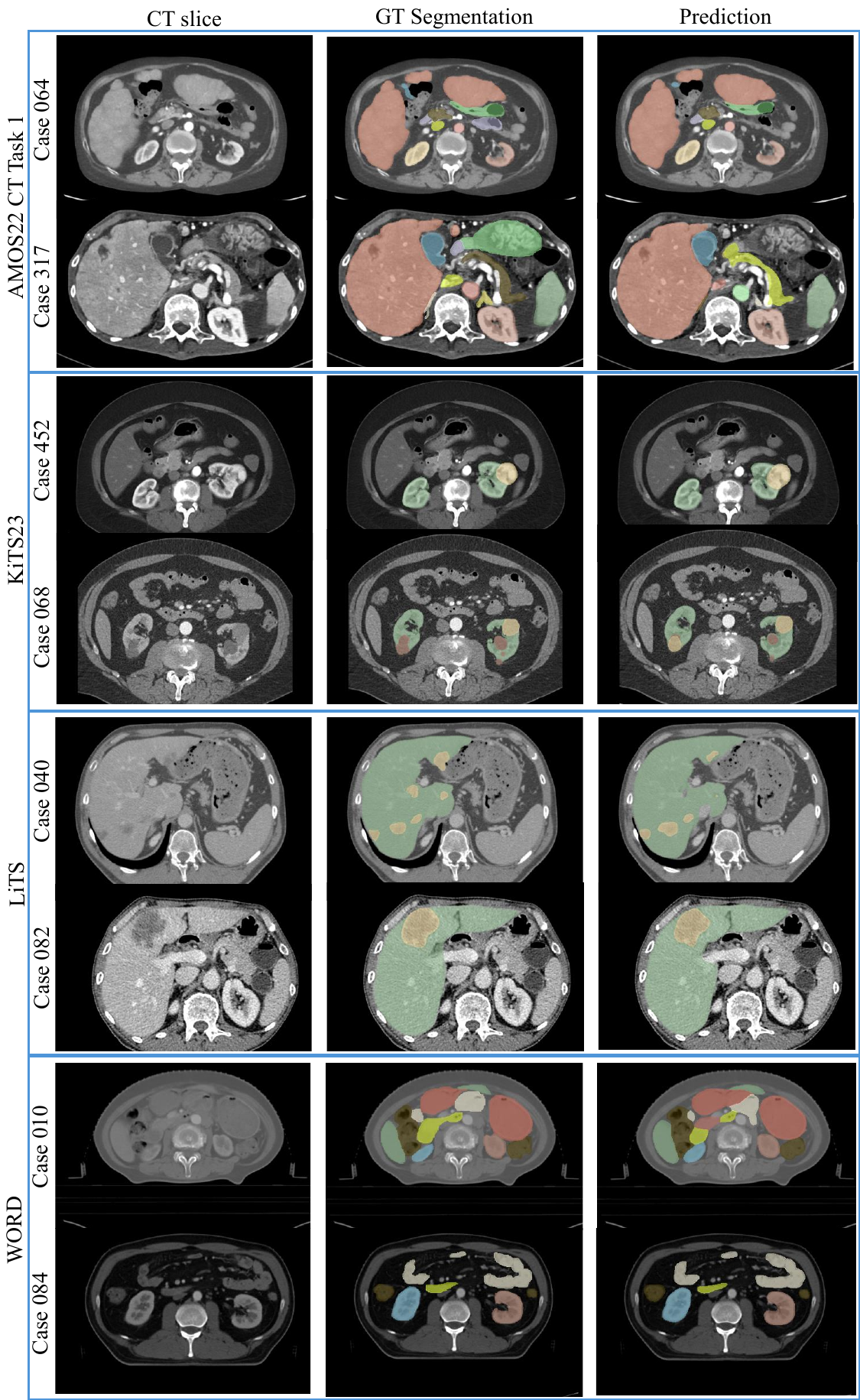}
    }
    \caption{Curated example semantic segmentation predictions of SPECTRE on the different datasets employed in this work. Predictions with good performance and with the worst performance are depicted. Window settings optimized for organs of interest.}
    \label{fig:segmentation_results_qualitative}
\end{figure*}

\cref{fig:segmentation_results_qualitative} shows qualitative masks for multiple datasets; they illustrate the same pattern as the numbers: large organs are clean and contiguous with the ground truth labels, while small tumors are present but slightly smoothed—consistent with predicting at 1/4~resolution and upsampling. We chose not to add an extra refinement head to keep the experiment honest.
\begin{table}[t]
    \centering
    \caption{Segmentation results (Dice) over the last 3 datasets. TS is TotalSegmentator.}
    \resizebox{0.9\columnwidth}{!}{%
        \begin{tabular}{l|l|ccc}
            \toprule
            & \multirow{2}{*}{\emph{Method}}  & \multicolumn{3}{c}{\emph{Dice (\%) $\uparrow$}} \\
            &    & TSv1-Full & TSv2-Full & TSv2-Merlin \\
            \midrule
            \multirow{4}{*}{\rotatebox[origin=c]{90}{Conv.}} 
            & SuPreM (U-Net)       & -     & 86.95 & -     \\
            & CT-FM (Res.U-Net)    & -     & 89.81 & 90.17 \\
            & Merlin (Res.U-Net)   & -     & -     & 86.20 \\
            & nnU-Net              & 85.22 & -     & -     \\
            \midrule
            \multirow{2}{*}{\rotatebox[origin=c]{90}{Trans.}}
            & SAM-Med3D (1 click) & 84.68  & - & - \\
            & SAM-Med3D (10 clicks) & 87.59  & - & - \\
            \midrule
            & SPECTRE (SEoMT) & 87.34 & 88.85 & 87.29 \\
            \bottomrule
        \end{tabular}%
    }
    \label{tab:segmentation_revisited}
\end{table}

\subsubsection{Additional Comparison on TotalSegmentator Benchmarks}
To further strengthen the segmentation study beyond KiTS23, LiTS, and WORD, we additionally evaluate SPECTRE on the three TotalSegmentator-based benchmarks reported in \cref{tab:segmentation_revisited}: \emph{TSv1-Full}, \emph{TSv2-Full}, and the more distribution-aligned \emph{TSv2-Merlin} subset. These experiments are included to assess whether the encoder-only SEoMT formulation remains competitive on broader anatomical segmentation tasks and to compare against recent CT foundation models that were explicitly designed for segmentation. In particular, we compare against SuPreM, CT-FM, and Merlin, and we additionally include SAM-Med3D as a strong interactive transformer baseline where available.

The results in \cref{tab:segmentation_revisited} show that SPECTRE with SEoMT remains consistently competitive across all three benchmarks. On \emph{TSv1-Full}, SPECTRE achieves a Dice score of $87.34\%$, improving over the conventional nnU-Net baseline ($85.22\%$) and also exceeding the one-click SAM-Med3D result ($84.68\%$). The ten-click SAM-Med3D result is slightly higher ($87.59\%$), but this comes at the cost of interactive prompting, whereas SPECTRE operates fully automatically in a feed-forward manner. On \emph{TSv2-Full}, SPECTRE reaches $88.85\%$, outperforming SuPreM ($86.95\%$) while trailing CT-FM ($89.81\%$). On the \emph{TSv2-Merlin} subset, SPECTRE obtains $87.29\%$, again remaining competitive and improving over Merlin, though CT-FM achieves the strongest score ($90.17\%$).

Overall, these results support two conclusions. First, the proposed encoder-only adaptation is not limited to kidney or lesion segmentation, but transfers well to large-scale multi-structure CT benchmarks. Second, although SEoMT is not intended as an aggressively optimized decoder for state-of-the-art segmentation, it provides strong performance with minimal decoder-specific bias and therefore offers a more direct probe of encoder feature quality. We therefore position these experiments primarily as evidence that the learned SPECTRE representation is broadly useful for segmentation, rather than as a claim that SEoMT is the final or optimal decoder for 3D CT. Future work should investigate stronger task-specific 3D decoders built on top of the same pretrained encoder.

\begin{table}[t]
    \centering

        \centering
        \captionof{table}{Segmentation results (DSC, NSD) using different decoders. All fold 0 retrained.}
        \resizebox{0.9\columnwidth}{!}{%
            \begin{tabular}{l|cc|ccc}
                \toprule
                \multirow{2}{*}{\emph{Method}}  &  \multicolumn{2}{c|}{KiTS23} &  \multicolumn{3}{c}{KiTS23~($<$10 mm masses)}    \\
                & \multicolumn{1}{c}{\emph{Dice (\%) $\uparrow$}} & \multicolumn{1}{c|}{\emph{NSD $\uparrow$}} &
                  \multicolumn{1}{c}{\emph{Dice (\%) $\uparrow$}} & \multicolumn{1}{c}{\emph{NSD $\uparrow$}} & \emph{Detect Rate} (\%)\\
                \midrule
                \multicolumn{6}{l}{37.5k Training Steps (150 epochs $\times$ 250 iterations)}\\
                \midrule
                SPECTRE (Linear) & 84.11 & 0.929 & 3.27$\times 10 ^{-5}$ & 0.01 & 2.0\\
                SPECTRE (SEoMT) & 86.70  & 0.943 & 3.76 $\times 10 ^{-4}$ & 0.02 & 1.53\\
                SPECTRE (UNETR) & 87.53 & 0.945 & 1.34 $\times 10 ^{-3}$ & 0.07 & 11.12\\
                \midrule
                \multicolumn{6}{l}{75k Training Steps (300 epochs $\times$ 250 iterations)}\\
                \midrule
                SPECTRE (Linear) & 85.42 & 0.934 & 4.72 $\times 10 ^{-5}$ & 0.03 & 1.81 \\
                SPECTRE (SEoMT) & 87.13 & 0.948& 4.11 $\times 10 ^{-4}$& 0.03 & 5.36\\
                SPECTRE (UNETR) & 87.82 & 0.948 &1.71 $\times 10 ^{-3}$ & 0.08 & 15.12\\
                \midrule
                \multicolumn{6}{l}{250k Training Steps (1000 epochs $\times$ 250 iterations)}\\
                \midrule
                nnU-Net ResEnc L & 88.26& 0.954& 2.18 $\times 10 ^{-3}$  & 0.11 & 22.79\\
                \bottomrule
            \end{tabular}
            \label{tab:segmentation_decoders}
        }
        \label{tab:segmentation_results_ablations}
\end{table}

\subsubsection{Decoder Variants and Small-Structure Analysis}
A potential concern with the encoder-only SEoMT design is that its simplicity may understate the true segmentation potential of the pretrained SPECTRE features. To study this explicitly, we compare three decoder choices on KiTS23 in \cref{tab:segmentation_results_ablations}: (1)~a \emph{Linear} decoder, which projects the encoder features directly to class logits; (2)~the proposed \emph{SEoMT} decoder, which appends class queries and performs joint self-attention in the final transformer blocks; and (3)~a stronger \emph{UNETR}-style decoder that introduces a more conventional task-specific decoding pathway. All models are retrained on fold~0 under matched training budgets, and we additionally report performance on the particularly challenging subset of lesions smaller than $10$~mm.

The full KiTS23 results show a clear ranking across decoder complexity. Under the $37.5$k-step setting (150~epochs), Linear reaches $84.11\%$ Dice and $0.929$ NSD, SEoMT improves this to $86.70\%$ Dice and $0.943$ NSD, and UNETR further increases performance to $87.53\%$ Dice and $0.945$ NSD. The same ordering remains after extending training to $75$k steps (300~epochs), where Linear obtains $85.42\%$, SEoMT $87.13\%$, and UNETR $87.82\%$ Dice. This consistent progression indicates that the SPECTRE encoder exposes useful dense features and that stronger decoders can indeed extract additional segmentation performance from them.

The analysis on tiny masses is even more informative. For lesions smaller than $10$~mm, all models perform substantially worse than on the full benchmark, confirming that this regime is intrinsically difficult. Nevertheless, the same decoder trend persists: Linear yields near-zero Dice and NSD, SEoMT improves modestly, and UNETR provides the best small-structure sensitivity, increasing the detection rate from $1.53\%$ with SEoMT to $11.12\%$ at $37.5$k steps and to $15.12\%$ at $75$k steps. For reference, the much longer-trained nnU-Net ResEnc L baseline, optimized for $250$k steps, reaches a detection rate of $22.79\%$. These results indicate that the limitation on very small structures is not caused solely by the pretrained representation, but also by the decoding strategy and the training budget.

Importantly, SEoMT was designed to evaluate encoder feature quality with minimal decoder bias rather than to maximize segmentation performance at all costs. In our implementation, trilinear interpolation is applied to the $1/4$-resolution \emph{feature maps}, not to already discretized masks, which preserves more fine-grained spatial information despite the lightweight decoding path. Even so, \cref{tab:segmentation_results_ablations} shows that a stronger decoder such as UNETR is beneficial, especially for tiny lesions. We therefore view SEoMT as a clean and informative encoder-centric evaluation protocol, while the decoder ablation confirms that future work on SPECTRE should investigate more expressive 3D decoders when absolute downstream segmentation performance is the primary objective.

\subsection{Zero-Shot Text-to-Image Retrieval}\label{app:downstream/embeddings}
We finally report additional details on the zero-shot text–to–image retrieval experiments conducted in parallel to the downstream evaluations. Our goal is to align the protocol as closely as possible with prior work; all retrieval metrics and data splits follow the procedures described in CT-RATE~\cite{hamamci_foundation_2024} and MERLIN~\cite{blankemeier_merlin_2024}.

\subsubsection{Retrieval on CT-RATE validation cohort}
For CT-RATE, we evaluate retrieval performance using both the \emph{Impressions} and \emph{Findings} sections of each radiology report. Following the original setup, each report is treated as a single textual query, and we compute Recall@\{5, 10, 50, 100\} on the full validation set of $N = 1{,}564$ studies. Retrieval is based on cosine similarity in the shared image–text embedding space, and a query is counted as correct if the paired CT scan appears among the top-$K$ nearest neighbors. 

\begin{figure}
    \centering
    \includegraphics[width=0.9\linewidth]{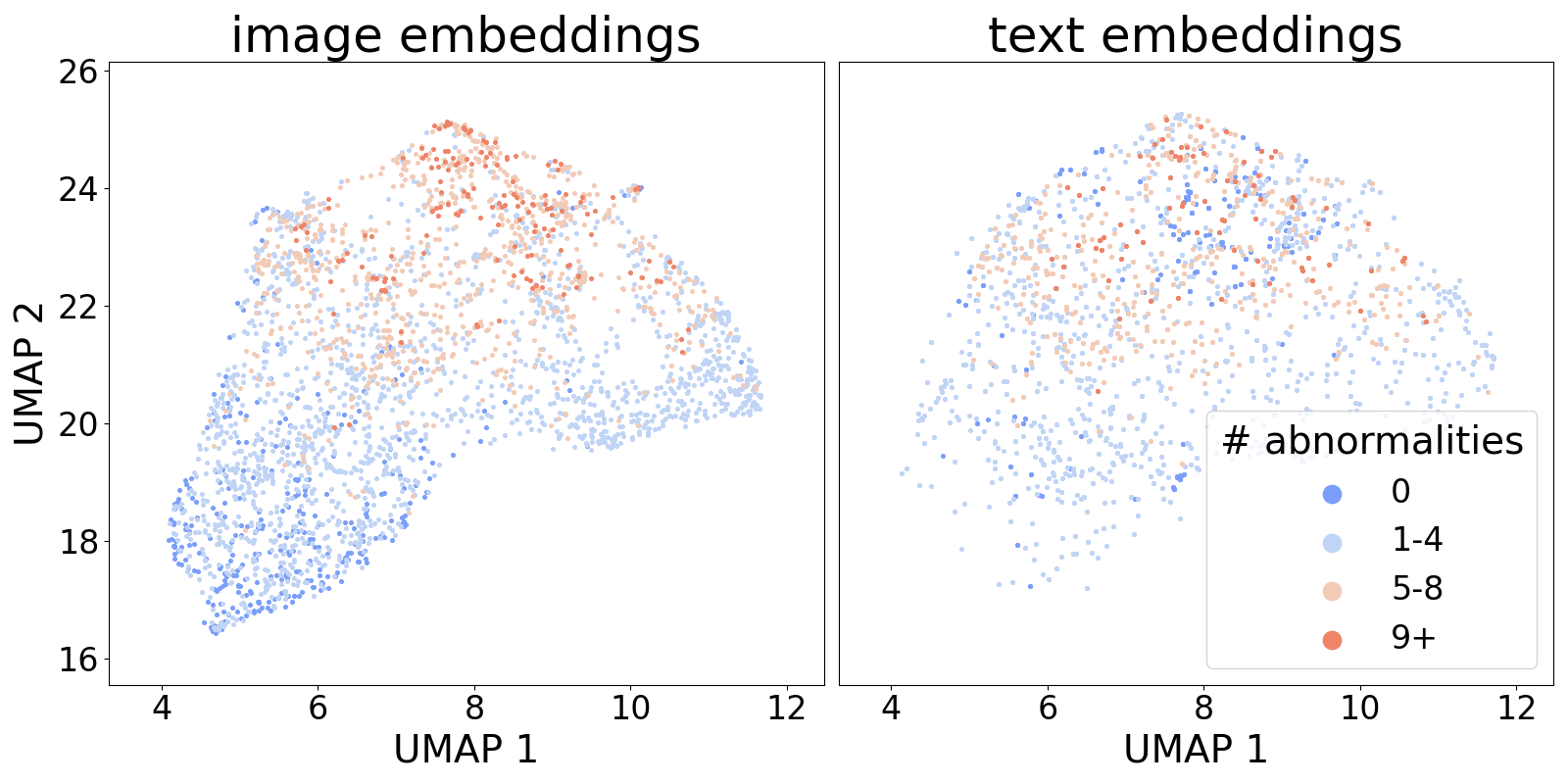}
    \caption{UMAP~\citep{mcinnes_umap_2018} visualization of image and text embeddings from the CT-RATE validation set~\citep{hamamci_foundation_2024}. Each point represents a sample categorized by the number of abnormalities noted in the corresponding radiology report.}
    \label{fig:umap}
\end{figure}

\cref{fig:umap} visualizes the joint embedding distribution of CT-RATE after UMAP projection, showing extensive cross-modal overlap and a smooth trajectory correlated with the total number of abnormalities described in the reports. This overlap indicates that the model learns a coherent shared latent space in which radiology images and their associated reports are embedded consistently, suggesting that the representations are largely modality-agnostic and capture clinically meaningful semantics rather than modality-specific artifacts. The continuous progression along the manifold with an increasing abnormality count further supports the notion that the embedding space encodes a graded representation of pathological severity or complexity.

We repeat the same experiment using MedSigLIP\footnote{\url{https://github.com/Google-Health/medsiglip}}, which forms the visual encoder of Google's MedGemma model~\cite{sellergren_medgemma_2025}. Retrieval performance is low, with Recall@\{5,~10,~50,~100\}=\{0.3,~0.7,~4.8,~8.2\}\%, only slightly above random chance. This limited performance can likely be attributed to the model’s restriction to 128 input tokens, which truncates longer radiological reports and prevents the model from accessing much of the available descriptive information.

All retrieval experiments are conducted using a fixed voxel spacing of $0.5\times0.5\times1.0$~mm. To assess the robustness of our model to variations in scan resolution and anisotropy, we also perform the CT-RATE experiment using each scan’s native spacing. We observe minimal performance change (-0.6\%~in~Recall@5), demonstrating that our model is largely insensitive to differences in voxel resolution and anisotropy, which is important for real-world clinical applicability.

\subsubsection{Retrieval on MERLIN test cohort}
For MERLIN, we mirror the evaluation strategy presented in the original paper. Retrieval is conducted separately for the \emph{Impressions} and the \emph{Findings} sections, and performance is quantified using Recall@\{1, 8\}. Rather than the full test set, MERLIN evaluates retrieval over sampling pools of fixed sizes~$N \in \{32, 64, 128\}$, each representing a different difficulty level. Cosine similarity is again used to rank image–text pairs, and correctness is assessed based on whether the paired CT volume is returned within the top-$K$ matches.

\begin{figure}
    \centering
    \includegraphics[width=0.9\linewidth]{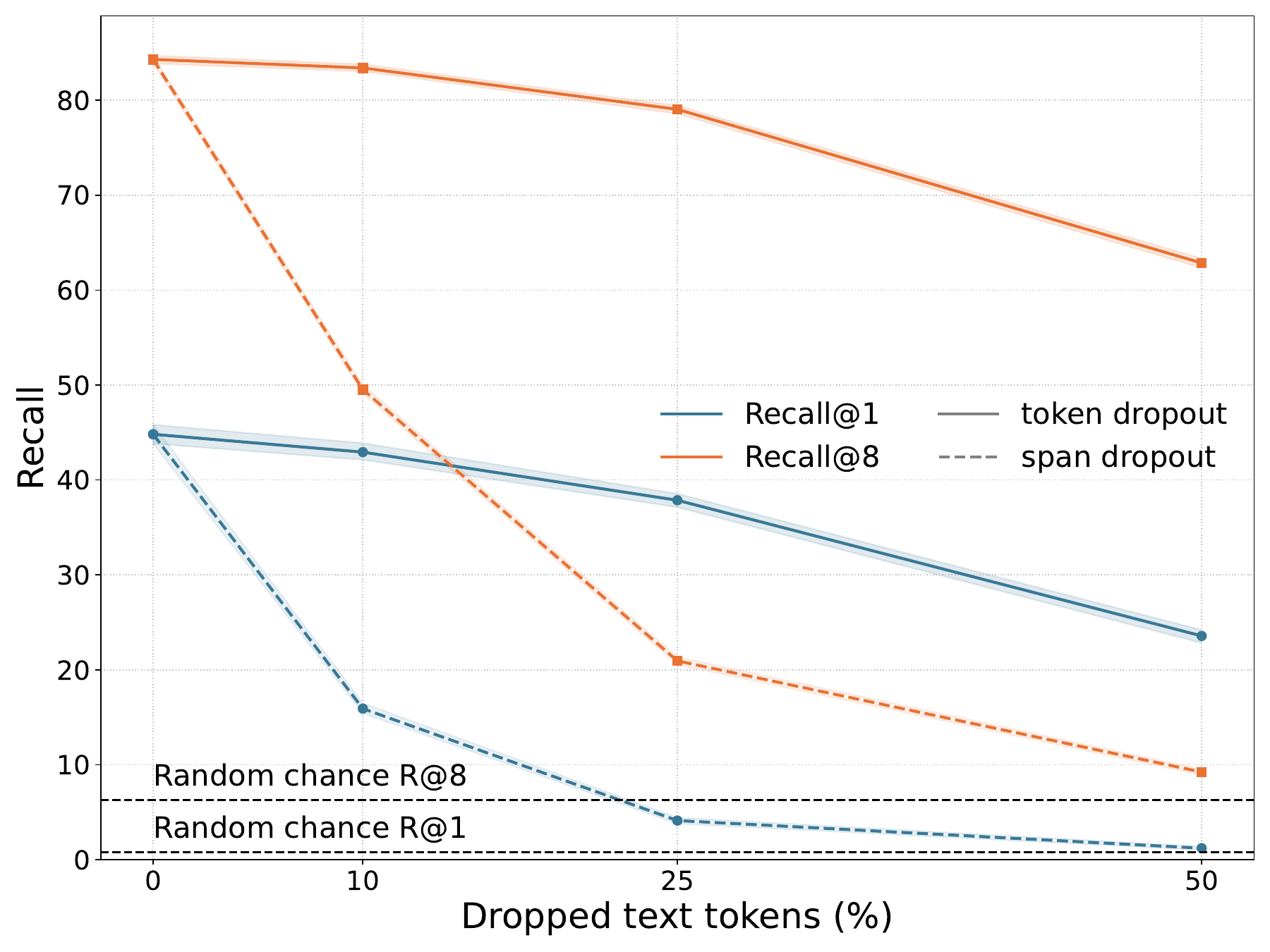}
    \caption{Impact of text dropout on retrieval performance. 
    }
    \label{fig:degradation token dropping}
\end{figure}

Medical reports are inherently noisy due to variability in clinicians’ writing styles, abbreviations, and selective reporting. To assess the robustness of our text-to-image retrieval model under such realistic noise conditions, we simulate report corruption in two complementary ways. First, we perform \emph{random token dropout}, which models inconsistencies in clinical phrasing. For instance, a report might mention ``tumor” rather than the more specific ``lung tumor,” reflecting incomplete or abbreviated descriptions. Second, we apply \emph{random span dropout}, where contiguous spans of 10–50~tokens are removed throughout the report to simulate missing observations or unrecorded findings. The results of these experiments are shown in \cref{fig:degradation token dropping}. As anticipated, model performance degrades more significantly under span dropout than token dropout, reflecting the greater impact of missing semantic content. Interestingly, performance remains relatively stable under token dropout: even when 25\% of all tokens are removed, the decline in Recall@\{1,~8\} both remain below 10\%. This demonstrates the robustness of the Qwen3 Embedding model with LoRA adapters in capturing medical language semantics, maintaining meaningful retrieval even when reports are partially incomplete. These findings highlight the model’s potential for real-world clinical applications, where reports are often imperfect or partially specified.

We further analyze the model performance with respect to report length by splitting the dataset into long reports (more than 500~tokens) and short reports (fewer than 500~tokens). We observe a notable difference in retrieval performance, with Recall@1=48.7\% for long reports compared to Recall@1=34.7\% for short reports. This suggests that the model effectively leverages the richer, more detailed information present in longer reports, allowing for more precise alignment with corresponding images. In contrast, shorter reports provide less context and fewer descriptive cues, which limits the model’s ability to establish strong associations. We note, however, that shorter reports often correspond to healthy subjects, where findings are minimal and reports tend to be more uniform, which could also contribute to the observed performance gap.

\subsection{Hardware}
All downstream and ablation experiments are performed on a single H100~GPU~(NVIDIA~Corp., CA, USA) containing 96~GB of GPU memory, hosted in a system equipped with an AMD 4th~Gen EPYC processor (18~cores, 36~threads) and 180~GB of system memory.

\newpage
{
    \small
    \bibliographystyle{ieeenat_fullname}
    \bibliography{references}
}

\end{document}